\documentclass[10pt,journal,compsoc]{IEEEtran}

\usepackage{psfrag}
\usepackage{mathrsfs}
\usepackage{yfonts}
\usepackage{enumitem}
\usepackage{breqn} %
\usepackage{hyperref}

\usepackage{graphicx}
\usepackage{caption,subcaption} 

\usepackage{algpseudocode}
\usepackage{color}
\usepackage{eucal}

\usepackage{amssymb}
\usepackage{amsthm}
\usepackage{amsfonts}
\usepackage{mathptmx}
\usepackage{eqnarray}

\usepackage{float}
\usepackage{booktabs}
\usepackage{multirow}
\usepackage{array}
\usepackage{algorithm}

\usepackage[numbers]{natbib}
\usepackage{lineno}
\usepackage{epstopdf}
\usepackage{microtype}
\usepackage{tcolorbox}
\usepackage[tikz]{bclogo}
\usepackage{lipsum}

\renewcommand{\figurename}{Fig.}
\usepackage{threeparttable}

\usepackage{siunitx} 

\usepackage{tikz}
\usetikzlibrary{shapes.geometric, arrows}
\usetikzlibrary{positioning}
\tikzstyle{startstop} = [rectangle, rounded corners, minimum width=3cm, minimum height=1cm,text centered, draw=black, fill=red!30]
\tikzstyle{io} = [trapezium, trapezium left angle=70, trapezium right angle=110, minimum width=3cm, minimum height=1cm, text centered, draw=black, fill=blue!30]
\tikzstyle{process} = [rectangle, minimum width=3cm, minimum height=1cm, text centered, draw=black, fill=orange!30]
\tikzstyle{decision} = [diamond,aspect=2, minimum width=3cm, minimum height=1cm, text centered, draw=black, fill=green!30]
\tikzstyle{arrow} = [thick,->,>=stealth]

\newcommand{\bfg}[1]{\mbox{\boldmath $#1$}}
\newcommand{\real}{\hbox{\rm I\kern-0.2emR}}
\newcommand{\ass}{\mathop{ {\bf A}}_{e=1}^{N_e} }
\newtheorem{remark}{Remark}[section]
\newtheorem{theorem}{Theorem}[section]

\newcommand{\argmin}{\mathop{\mathrm{argmin}}}

\DeclareMathAlphabet{\mathcal}{OMS}{cmsy}{m}{n}

\begin{document}

\title{
A Variational Bayesian Inference Theory of
Elasticity and Its Mixed Probabilistic Finite
Element Method for Inverse Deformation Solutions in
Any Dimension
\color{black}
}

\author{Chao~Wang
        and~Shaofan~Li
\IEEEcompsocitemizethanks{\IEEEcompsocthanksitem C.~Wang and S.~Li are with the Department
of Civil and Environmental Engineering, The University of California, Berkeley,
CA, 94720, USA.
E-mail: shaofan@berkeley.edu
\IEEEcompsocthanksitem C.~Wang is the first author. S.~Li is the corresponding author}
}


\IEEEtitleabstractindextext{%
\begin{abstract}
In this work, we have developed a variational Bayesian inference theory of 
elasticity, which is accomplished by using
a mixed Variational Bayesian inference
Finite Element Method (VBI-FEM) that can be used to solve the inverse 
deformation problems of continua.
In the proposed variational Bayesian inference theory of continuum mechanics,
the elastic strain energy is used as a prior in a Bayesian inference
network, which can intelligently recover the detailed continuum deformation mappings
with only given the information on the deformed and undeformed continuum
body shapes without
knowing the interior deformation and the precise actual boundary conditions, 
both traction as well as displacement boundary
conditions, and the actual material constitutive relation.
Moreover, we have implemented the related
finite element formulation in a computational probabilistic mechanics framework.
To numerically solve mixed variational problem,
we developed an operator splitting or staggered algorithm
that consists of
the finite element (FE) step and the Bayesian learning (BL) step as
an analogue of the well-known the Expectation-Maximization (EM) algorithm.
By solving the mixed probabilistic Galerkin variational problem,
we demonstrated that the proposed method is able to
inversely predict continuum deformation mappings
with strong discontinuity or fracture
without knowing the external load conditions.
The proposed method provides a robust machine intelligent solution 
for the long-sought-after inverse problem solution, 
which has been a major challenge
in structure failure forensic pattern analysis
in past several decades.
The proposed method may become a promising artificial intelligence-based inverse method for
solving general partial differential equations.
\end{abstract}

}

\maketitle

\IEEEpeerreviewmaketitle

\begin{figure}[ht]
\begin{center}
\includegraphics[width=0.9\columnwidth]{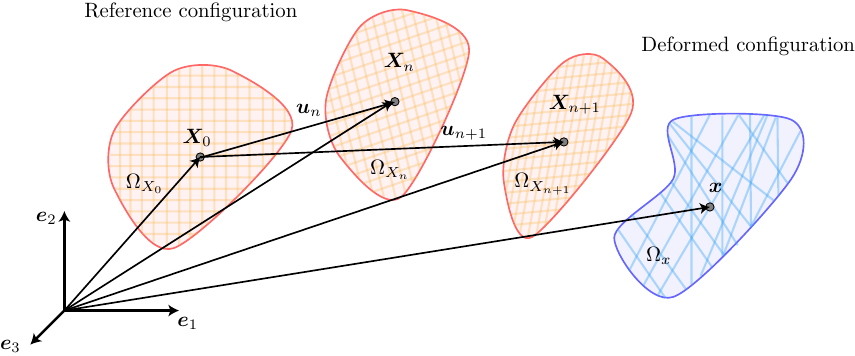}
\end{center}
\caption{Schematic illustration of the inverse solution of continuum deformation mapping recovery problem.}
\label{fig:deformation map recovery problem}
\end{figure}

\IEEEraisesectionheading{
\section{Introduction}}
In continuum solid mechanics, specifically in elasticity theory,
under external loading,
a continuum object deforms from the referential configuration $\Omega_X$
to a deformed configuration $\Omega_x$. This deformation process
can be described by a continuous function ${ \psi}: \Omega_X \to \Omega_x$,
which is called the deformation map or mapping (see Fig. \ref{fig:deformation map recovery problem}).
The fundamental problem of continuum mechanics is to find the
three-dimensional(3D) deformation
mapping under given boundary loading conditions.
Boundary value problem solutions in continuum mechanics have been well-developed including
analytical solutions as well as numerical solutions, such as
finite element solutions e.g. \cite{Liu2022,hughes2003,zienkiewicz2005finite}.

However, in many engineering applications, we do not know the precise boundary conditions.
In many cases, we may not even know the material properties of the deformed solids.
The only known information or data available
is the boundary contour shape
of the deformed solid,
which may be obtained by various image processing procedures.
Nevertheless, we would like to know
the precise three-dimensional deformation field of the deformed continuum solid,
so that we
can find the strain field
and maybe even the stress fields of the continuum.
This is the fundamental inverse problem of continuum mechanics.

In general, we may have two mesh data in both the
initial (referential) configuration and the final (current)
configuration, but the coordinates of the
data points in the initial and the final configurations have
no correlation.
The inverse problem in continuum mechanics is
to find the continuum deformation
map or correlation between these two sets of data
of mesh points
without knowing the actual boundary conditions
such as the loading conditions as well as the precise material properties
\cite{bonnet2005inverse}.
This is because in many engineering applications, that information is either
unknown or incomplete.
For instance, after a traffic collision,
one may only have the image of the car wreckage configuration
without knowing the details of the collision process,
and thus it is impossible to find the deformation map of
the collision cars by using the forward analysis or calculation.
On the other hand, in a traffic accident
forensic analysis, we need the residual deformation field
to infer the pre-crash information,
such as impact speed, offset, and angle (see \cite{chen2021deep,Xie2022}).
Despite the fact that the forward analysis procedure
of solid mechanics has been well-established,
its inverse problem solution or prediction remains as a
challenging problem \cite{bui2007fracture, turco2017tools}.
Theoretical and numerical techniques of the inverse problem solution
have been developed over the years. For instance,
Alves et al. \cite{alves1999inverse} proposed a theoretical
formulation to characterize the inverse scattering
for elastic plane cracks.
Ballard et al. \cite{ballard1994inversion} discussed the reconstruction of the stress field from the
measurements stress on the boundary.
Weglein et al. \cite{weglein2003inverse} proposed a mathematical-physics framework
for the inverse scattering series.
Guzina et al. \cite{guzina2004topological} established an alternative analytical
framework for the inverse scattering of elastic waves.

With the explosive developments and applications of artificial intelligence and data science,
machine learning-based approaches have received much attention.
Among all ML approaches, deep neural networks (DNN) lead to excellent performance
in a large number of fields, such as computer vision \cite{voulodimos2018deep}
and natural language processing \cite{hochreiter1997long}.
Besides the achievements in computer science, DNN is also applied to the inverse problem solutions
in mechanics.
Chen et al. \cite{chen2019application} proposed a novel deep-learning inverse solution
to determine the loading conditions of the shell structure. After training the DNN,
the loading can be identified by the information on the plastic deformation of a shell structure.
Ni et al. \cite{ni2021deep} discussed the application of DNN
in identifying the elastic modulus based on the measured strain field.
By collecting typical representative of shear modulus, the trained DNN
can learn the mapping between stress and strain with high accuracy.
A deep energy approach \cite{samaniego2020energy, nguyen2021parametric, li2021physics}
was also proposed to identify the displacement field by structural energy-informed DNN.
In addition to feed-forward neural networks, image-based convolutional neural network (CNN)
\cite{goodfellow2016deep} was also introduced in computational mechanics.
Gao et al. \cite{gao2022physics} proposed a physics-informed graph neural Galerkin
network which is a unified method for both forward and inverse problems.
Unlike traditional physics-informed neural networks (PINNs) which suffer from unsatisfactory
scalability and hard boundary enforcement, piecewise polynomial basis function was applied
to interpolate the search space.

Recently,
the Gaussian Mixture Model (GMM) based non-rigid registration methods in computer vision,
 e.g. \cite{reynolds2009gaussian,myronenko2010point, nguyen2015multiple,
 ma2015non, wang2021quantification, xie2022generalized, xie2023bayesian},
 were applied in the inverse analysis of deformation matching.
In this field, the coherent point drift (CPD) \cite{myronenko2010point}
approach is a popular method, and it has several variations.
Using the GMM technique, the registration problem in CPD is formulated
as a maximum likelihood (ML)
estimation problem. GMM centroids are used to represent one point set,
and by moving coherently,
another point set is fitted to the first one's centroids.
However, the current non-rigid registration method is still limited
to the one-to-one matching process of
data points on the surface of the peripheral or layout boundary of a continuum object, and
it is not designed to inversely recover the three-dimensional
continuum deformation mapping in engineering applications, in which
many deformations may be discontinuous or have strong discontinuities.

 In this work, we have developed
 a variational Bayesian inference finite element method (VBI-FEM)
 based on a mixed probabilistic variational principle of
 Bayesian inference, which takes advantage of the powerful FEM solution
 and incorporates the GMM-based machine learning edge
 into a three-dimensional FEM framework,
 to inversely recover three-dimensional deformation
 mappings or patterns
 {
with any uniform surface strain fields.
 }

 Following the introduction, in Section \ref{sec: Inverse problem statement},
 we first briefly
 introduce the inverse deformation statement and its relationship with the forward problem.
 Then we discuss the probabilistic variational formulation and the corresponding operator
 splitting solution algorithm consisting
 the Finite Element (FE) solution step and the Bayesian Learning (BL) solution step
 in Section \ref{sec: Entropic variational theory}.
 The details of the formulation and implementation of VBI-FEM are discussed
 in Section \ref{sec: VE-FEM formulation and implementation}.
 In Section \ref{sec:example}, the proposed VBI-FEM is demonstrated and validated by several
 numerical examples. Finally, the conclusions are drawn in Section \ref{sec: Conclusion}.
\begin{figure}[ht]
\centering
  \includegraphics[width=\columnwidth]{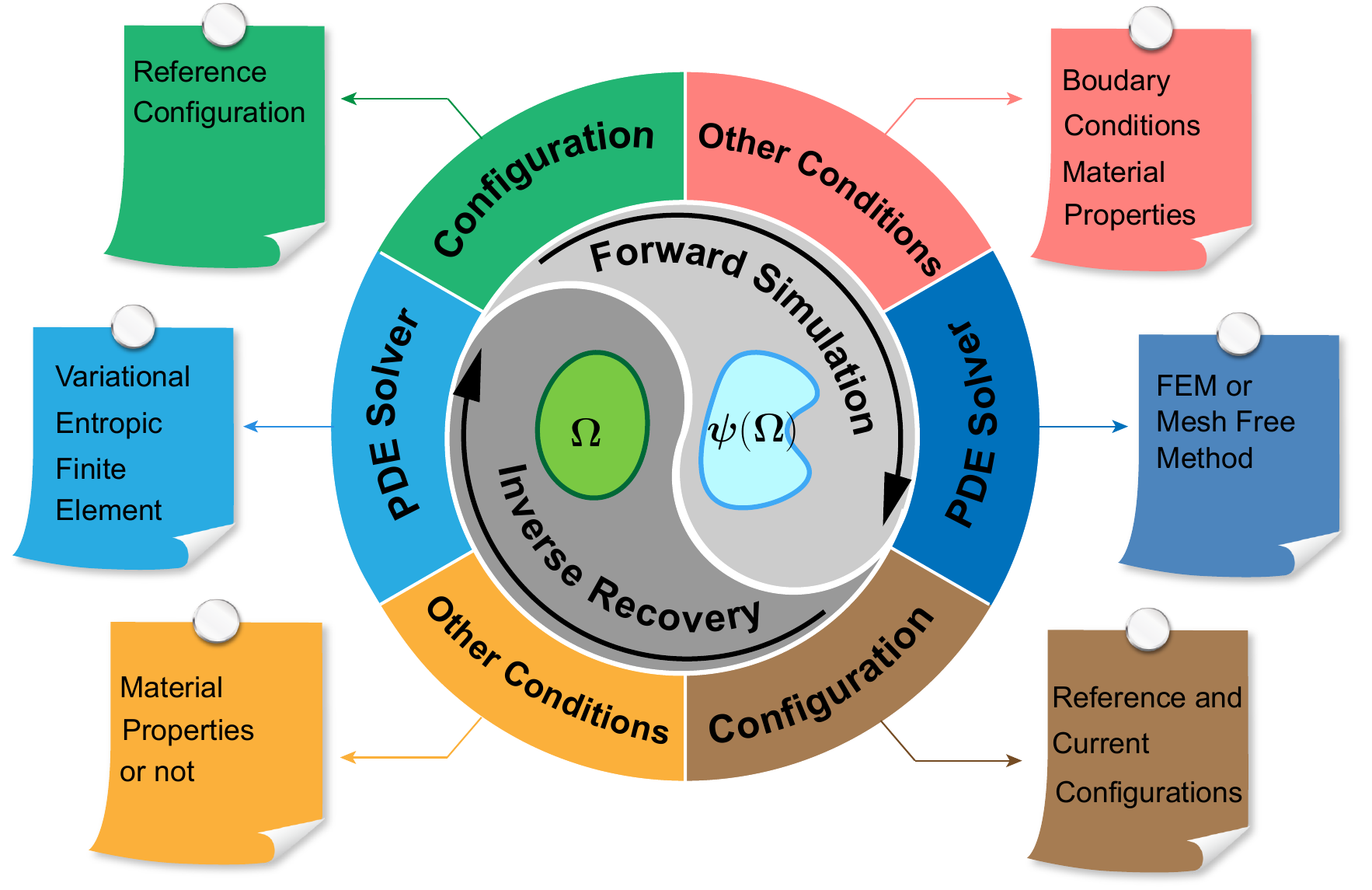}
\caption{ Comparison between the
forward boundary-value problem solution and the inverse problem solution.}
\label{fig:BVP vs inverse problem}
\end{figure}

\section{Inverse identification of continuum deformation mappings}
\label{sec: Inverse problem statement}
Before we proceed,
we first state the inverse problem of continuum deformation mapping.
Let $\Omega_0$ and ${ \psi} (\Omega_0) =\Omega$  be the referential
configuration and current configuration in $\mathbb{R}^{3}$ respectively as
shown in Fig. \ref{fig:deformation map recovery problem}.
Points ${\bf X} \in \Omega_{0}$ are called material points.
Then the deformation mapping $ \psi$ is defined
as a mapping: ${ \psi}: \Omega \to  { \psi}(\Omega)$,
which assigns a new position ${\bf x}$ to each material point ${\bf X}$ in $\Omega_{0}$ as,
\begin{equation}
    {\bf x} = { \psi} ({\bf X}).
\end{equation}
Therefore, we may denote
the reference and current configurations $\Omega$
and ${ \psi}(\Omega)$ as $\Omega_X$ and $\Omega_x$ respectively.

Given boundary conditions and a constitutive model,
for infinitesimal deformation,
the conventional continuum mechanics can determine the deformation mapping
or the associated displacement field
by solving the following
boundary-value problem (BVP) that can be summarized below:

\begin{tcolorbox}
\textit{Infinitesimal Deformation Problem (Strong form).}\\
Find ${ \psi}: \Omega  \longrightarrow { \psi}(\Omega) \subset \mathbb{R}^{n}$ and
the corresponding displacement field ${\bf u} = { \psi} ({\bf X}) -{\bf X}$, satisfying
\begin{equation}
\begin{array}{llll}
&{\rm{div}}
[ \sigma({\bf u})] + {\rho} {\mathbf b}
= \mathbf{0}  ~~~ & \textrm{in} ~~~ &\Omega_X~,\\
&\mathbf{u} = \bar{\bf{u}} ~~~ &\textrm{on} ~~~ &\Gamma_{u}~,\\
&{\sigma} \cdot {\bf n} = \bar{\bf{t}} ~~~ &\textrm{on} ~~~ &\Gamma_{t}~,
\end{array}
\label{eq:strong form}
\end{equation}
where $\rho$ is the material density, ${\bf b}$ is the body force per unit mass;
$\overline{\bf{t}}$ is the prescribed traction on the traction boundary, and
$\overline{\bf{u}}$ is the imposed displacement on the displacement boundary.
The stress ${\sigma}={\sigma}(\bf{u})$
is determined by the given constitutive model, and the strain field is defined as
${\varepsilon}(\bf{u})=Sym \nabla \otimes (\bf{u})$.
\end{tcolorbox}
\medskip
Note that in Eq. (\ref{eq:strong form}) $\Omega_X~ \subset \mathbb{R}^{n}$ (for $n = 1,2,3$)
is a bounded domain, and $\partial \Omega_X  = \Gamma_{u} \cup \Gamma_{t}$, in which
$\Gamma_{u}$ is the prescribed displacement boundary and $\Gamma_{t}$
is the prescribed traction boundary, and
\begin{equation}
    \Gamma_{u} \cap \Gamma_{t} = \varnothing ~.
\label{eq:boudnary condition}
\end{equation}
The deformation map ${\psi}$ can be determined from the displacement field as
\begin{equation}
   {\psi} ({\bf u}) = {\bf x}= {\bf X} + {\bf u},
\end{equation}
which is usually a one-to-one corresponding mapping.

By assuming a linear elastic isotropic material, the constitutive relationship
 between the Cauchy stress  ${\sigma}$ and infinitesimal strain
 ${\varepsilon}$ can be expressed as
\begin{equation}
   {\sigma} = {\frac{\partial W}{\partial \varepsilon}}
    = \lambda \textrm{tr}[\varepsilon]
    {\bf I} +  2\mu \varepsilon,
\end{equation}
where $W=W({\varepsilon})$ is the strain energy density of a model solid,
while $\lambda$ and $\mu$ are Lam\'e constants. Note that the model solid
may not be the real solid whose deformation we intend to recover.

In the boundary value problem (BVP) of the Navier equation of linear elasticity, displacement
field ${\bf u}({\bf X})$ is the only unknown.
To solve the linear elasticity boundary value problem stated in
Eq. (\ref{eq:strong form}), the boundary conditions, external loads,
material constants and reference configuration must be given. These conditions
are the inputs to
a chosen partial differential equation (PDE) solver, e.g. finite element method (FEM).

The inverse problem referring to the above forward solution is
that we aim to solve the above problem without knowing the boundary conditions including
the precise information on $\Gamma_u$ and $\Gamma_t$ as well as
the precise material constitutive relations. The only information that we have is the
referential configuration of the undeformed solid body, or the shape of the undeformed
body, $\Omega_{X}$ and the deformed configuration of, or the shape of, the solid body $\Omega_{x}$.
Mathematically, we seek to find a one-to-one deformation mapping such that
\begin{equation}
{\bf x} = {\psi} ({\bf X}),~~\forall~~{\bf x}
\in  \Omega_x,~{\rm and}~\forall~{\bf X} \in \Omega_X ~.
\end{equation}

In passing we note that $\Omega_x$ needs not to be continuous everywhere, and
it may contain strong discontinuities, such as cracks.
Since the displacement field determines the deformation map,
we need to find the displacement field in this inverse problem setting,
in which the external loading conditions are unknown,
and the displacement boundary conditions are also unknown.
The only available information is
the general shape or layouts of the reference configuration and current configuration, as
described in Fig. \ref{fig:BVP vs inverse problem}.
The comparison between the linear elastic BVP and its inverse
deformation map recovery problem is illustrated in Fig. \ref{fig:BVP vs inverse problem}.
In applications, this type of problem often occurs in material and structure forensic analysis,
structure failure analysis,
as well as general deformation measurements in complex engineering settings such as nanoscale
deformation measurements, determining plastic deformation of a crashed vehicle,
determine 3D printed product shape distortion, and
 hazard environment deformation probe, among many others.
In the above situations, we only have the image or scanned data of the deformed
structure parts, but we do not know the displacement field for the given material
points, which is usually impossible to measure.

\section{A mixed entropic variational principle of probabilistic elastic continuum}
\label{sec: Entropic variational theory}

In continuum mechanics, almost all finite element method formulations are based on energetic
variational principles.
In this work, we developed a finite element method based on a probabilistic
entropic variational principle
to solve the inverse problem of continuum mechanics.
To do so, we first consider
a generative model \cite{hastie2009elements} to describe the formulation of ${\bf x}$ from ${\bf X}$.
As a starting point, we consider that the reference configuration and
current configuration are represented by two different meshes or two different sets of points
that have no one-to-one correspondence
as shown in Fig. \ref{fig:deformation map recovery problem}.
Using the terminology in computer vision,
we may call the points $\mathbf{X}$ in the reference configuration as the centroids
of a Gaussian Mixture Model (GMM) and the points $\mathbf{x}$ in the current configuration
 as the data points, which are usually the scan data or image data.
Under this circumstance, we can move the centroid points $\mathbf{X}$ in the reference configuration
towards the data points $\mathbf{x}$ in the current configuration to recover the deformation map ${{\psi}_n}$.
Similar to the updated Lagrangian method,
the recovery displacement ${\bf u}_n({\bf X})$ in the deformation recovery process is defined as
\begin{equation}
    {\bf u}_n({\bf X})  = {\bf X}_n - {\bf X}_0 ~~\to~~ {\psi}_n ({\bf {X}})= {\bf X}_0 + {\bf u}_n
\end{equation}
where ${\bf X}_0={\bf X}$ is the initial position for the material point ${\bf X}$
in the reference configuration,
and ${\bf  X}_n$ is the updated position for the same material point after $n$th iteration.
In computation,
all the quantities are computed in the moving mesh in the updated configuration.

It may be noted during the updated Lagrangian solution
that the determination of which configuration is under
move and which is kept fixed is totally arbitrary.
Therefore, it is feasible to move the current configuration
${\Omega_{x}}$ toward the reference configuration,
$\Omega_{X}$,
to recover the inverse of deformation mapping ${{\psi}^{-1}_n}$.
We can let ${\bf x}_0 ={\bf x}$ and
the recovered displacement field is defined as
\begin{equation}
{\bf u}_n ({\bf x})= {\bf x}_0 - {\bf x}_n~~\to~~{\psi}^{-1}_n ({\bf x})
= {\bf x}_0 - {\bf u}_n ({\bf x})~.
\end{equation}
 This inverse recovery setting would be further discussed in subsection
 \ref{section: cyliner example}.
 Again, we stress that ${\bf x}_0={\bf x}$ and ${\bf X}_0={\bf X}$
 are not the same material point.

Nevertheless, in the following presentation,
we assume that the material point ${\bf X} \in \Omega_X$
moves towards the current configuration.
{
During the alignment, the possible displacement field is represented
by $\tilde{\bf u}$ and the intermediate possible displacement field after n-th
iteration step is represented as $\tilde{\bf u}_n$.
Therefore, the corresponding possible material point $\tilde{\bf X}_n$
can be represented as $\tilde{\bf X}_n = {\bf X} + \tilde{\bf u}_n$.
}
Then for a specific real data point $\mathbf{x} \in \Omega_x$,
a generative model can be established in
a two-step procedure. First, a latent variable ${{\tilde{\bf X}_n}} \in \Omega_{X_n}$
is generated, and
then depending on the value of the probability distribution parameter $\Delta$,
a data point ${\bf x} \in \Omega_x$ can be related with the variable ${\tilde{{\bf X}}_n}$
 by the selected Gaussian distribution centered at ${{\tilde{\bf X}}_n}$ as follows,
\begin{equation}
p(\mathbf{x}, {\tilde{{\bf X}}_n}) =
    p({\tilde{{\bf X}}_n})
    p(\mathbf{x} | {{\tilde{{\bf X}}_n}})~.
\end{equation}
Then, by the law of total probability,
the total probability of the specific data point ${\bf x}$
is given by
\begin{equation}
    p(\mathbf{x}| {\tilde{\bf u}_n, \Sigma_n}) = \int_{\Omega_{X}}
    p({\tilde{\bf X}_n | \tilde{\bf u}_n, \Sigma_n})
    p(\mathbf{x} | {\tilde{{\bf X}}_n;\tilde{\bf u}_n, \Sigma_n})d\Omega_{X},
    \label{eq:integral1}
\end{equation}
where $p({\tilde{\bf X}_n | \tilde{\bf u}_n, \Sigma_n})$ is the prior distribution of the material point ${\bf X}$.
After applying the corresponding displacement
${\bf u}_{n}$, the Gaussian distribution $ p(\mathbf{x} | {\tilde{{\bf X}}_n;\tilde{\bf u}_n, \Sigma_n})$ is given as
\begin{equation}
     p(\mathbf{x} | {\tilde{{\bf X}}_n;\tilde{\bf u}_n, \Sigma_n}) = \frac{1}{\left(2 \pi \Sigma_n^{2}\right)^{D / 2}}
 \exp \Bigl( {-\frac{\left\| {\tilde{\bf X}_n} - {\bf x} \right\| ^2}{2 \Sigma_n^{2}}}
 \Bigr).
\end{equation}

In order to distinguish the Cauchy stress from the Gaussian distribution variance,
the variance is denoted by the capital letter $\Sigma$.
Then the log-likelihood of the probability of all data points $\mathbf{x}$ can be expressed as
\begin{eqnarray}
    &&\int_{\Omega_x}\log(p(\mathbf{x}| {\tilde{\bf u}_n, \Sigma_n}))
    d\Omega_x \\
    &=& \int_{\Omega_x}
    \log\Bigl(\int_{\Omega_{X_n}}
    p({\tilde{\bf X}_n | \tilde{\bf u}_n, \Sigma_n})
    p(\mathbf{x} | {\tilde{{\bf X}}_n;\tilde{\bf u}_n, \Sigma_n})d\Omega_{X_n}\Bigr)d\Omega_x.
    \nonumber
    \label{eq:integral2}
\end{eqnarray}

In continuum mechanics e.g. \cite{Dow1998}, based on the principle of minimum potential energy,
the variational form of a continuum mechanics problem with infinitesimal deformation
can be stated as looking for a deformation solution ${\bf u}^{\star} ({\bf x})$ such that
\begin{equation}
\Pi ({\bf u}^{\star}) =
  \inf_{{\bf u} \in \mathcal{U}} \Pi ({\bf u})
\end{equation}
where $\mathcal{U}$ is the space of all the kinematic admissible displacements,
and
\begin{equation}
\Pi({\bf u})= \int_{\Omega}
W(\bfg{\varepsilon}({\bf u})) d \Omega + \Pi_{ext}({\bf u}) ,
\end{equation}
where $W(\bfg{\varepsilon}({\bf u}))$ is the strain energy state in a continuum,
and $\Pi({\bf u})$ is the total potential energy and $\Pi_{ext}({\bf u})$
is the potential energy of the external loading.
To solve this infinitesimal mechanical problem, one should find ${\bf u}^{\star}: \Omega \rightarrow \mathbb{R}^{D}$ which not only satisfies the essential boundary conditions ${\bf u}=\overline{{\bf u}}$ on $\Gamma_u$, but also makes the potential energy minimum.

Similarly, the strain energy state $W(\bfg{\varepsilon}({\bf u}))$ can be combined
with the data set probability $ p(\mathbf{x} | {\tilde{{\bf X}}_n;\tilde{\bf u}_n, \Sigma_n})$
to form the following deformation recovery problem.

\medskip

\begin{tcolorbox}
\textit{Deformation Recovery Inverse Problem.}
Find ${\tilde{\bf u}}: \Omega \longrightarrow \mathbb{R}^{D}$  and variance $\Sigma^2$,
such that it minimizes the probabilistic elastic entropy potential energy
\begin{equation}
({\bf u}^{\star}, {\Sigma^\star}) = \argmin\limits_{{\tilde{\bf u}}, \Sigma
\in \mathcal{V}\oplus \mathbb{R}^+}
{\tilde{\Pi}}({\tilde{\bf u}},\Sigma)
\end{equation}
where the probabilistic potential energy is given as
\begin{eqnarray}
&&{\tilde{\Pi}} ( {\tilde{\bf u}}, \Sigma) :=
\nonumber
\\
&-&
\int_{\Omega_x}
    \log\Bigl(\int_{\Omega_{X}}
    p({\tilde{\bf X} | \tilde{\bf u}, \Sigma})
    p(\mathbf{x} | {\tilde{{\bf X}};\tilde{\bf u}, \Sigma})d\Omega_{X}\Bigr)d\Omega_x
    \nonumber
    \\
    &+& \gamma \int_{\Omega_X} \phi ({\tilde{\bf u}}) d \Omega_X,
   \label{eq:entropy2}
\end{eqnarray}
where
$
\phi ({\tilde{\bf u}}) = {\frac{1}{2}} {\tilde{\bf u}}^2
$
is a regularization function that stabilizes computations, ${\bf \tilde{X} =
X + \tilde{u}}$
and $\gamma$ is a hyperparameter.
We choose the probability of the strain energy state $W({\tilde{\bf u}})$
and the probability of the $L_2$ error norm between the current positions
of material points and their referential position as the likelihood function, i.e.
\begin{eqnarray}
   p({\tilde{\bf X}| \tilde{\bf u},\Sigma}) &
    = \exp (-\beta W({\tilde{\bf u}}) ) \tilde{p}_0({\bf X}),
\end{eqnarray}
where $\tilde{p}_0 ({\bf X})$ is a uniform distribution, but
\begin{equation}
{
\tilde{p}_0 ({\bf X})  = p_0 ({\bf X})/ A
}
\end{equation}
here the constant $A$ is determined as
\[
A= \int_{\Omega_X} \exp (-\beta W({\tilde{\bf u}}) ) p_0({\bf X})
d \Omega_X,
\]
and $p_0 ({\bf X}) $is the original material probability distribution
 and is also assumed as a uniform
probability distribution in the referential
configuration.
Moreover, in this work, the conditional probability
 $p ({\bf x}| \tilde{\bf X}; \tilde{\bf u}, {\bfg \Sigma})$
is chosen as the following form of the Gaussian mixture,
\color{black}
\begin{equation}
     p(\mathbf{x} | {\tilde{{\bf X}};\tilde{\bf u}, \Sigma})
     = \frac{1}{\left(2 \pi \Sigma_n^{2}\right)^{D / 2}}
 \exp \Bigl( {-\frac{\left\| {\tilde{\bf X}} - {\bf x} \right\| ^2}{2 \Sigma_n^{2}}}
 \Bigr).
 \label{eq:G-Mixture}
\end{equation}

Here $D \ge 1$ can be any spatial dimension, and in this paper,
we specifically demonstrate the
algorithm for $D = 2, 3$.
\end{tcolorbox}

\medskip

\begin{remark}
\label{remark theory}
{\bf 1.}
One can see that the above probabilistic elastic potential entropy
has two sources: (a) The material-based probabilistic entropy,
and (b) the deformation-based probabilistic entropy.
In this work, we choose the material-based probability distribution
as a uniform probability distribution, which means that we do not consider
the effects of material heterogeneity, which will be studied in
the late sequence of this work.
{\bf 2.}
Hyperparameters $\gamma$ and $\beta$ both represent the amount
of regularization in the entropy potential.
$\beta$ controls the magnitude of the regularization function directly.
$\gamma$ reflects the relationship between the strain energy density
and material probability distribution.
Large values of $\gamma$ and $\beta$ typically produce smoother results but underfit the model.
The two hyperparameters can be tuned until finding the best matching result.
{\bf 3.}
When the dimension of the spatial space is 2 or 3, the normal distribution typically
manifests as a multivariate normal distribution, allowing for variations in variance along different axes.
Consequently, this feature can provide a representation or description of highly non-uniform
deformation fields in heterogeneous materials such as non-uniform surface strains.
In the present work, to simplify the mathematical model,
we made the simplifying assumption that the Gaussian variance is uniform in all directions.
Hence, as a result, we are limited to recovering deformation models characterized
by constant surface strain fields. For this reason, the present inverse solution
is not able to distinguish the solutions that have the same deformed shape with different
surface strains.
More complex cases will be studied in future work.
\end{remark}

The essence of the VBI-FEM is to minimize the probabilistic elastic potential
entropy by using the finite element method.
Analogous to the Expectation-Maximization (EM) algorithm
\cite{dempster1977maximum, bishop1995neural},
the mixed functional in Eq. (\ref{eq:entropy2})
can also be solved by using a staggered computational algorithm that is coined
as the Finite Element-Bayesian Learning (FE-BL) operator splitting algorithm.

{
Considering the first term in $\tilde{\Pi} ( {\tilde{\bf u}}, \Sigma)$
and apply the evidence lower bound ($\mathbb{ELBO}$) identity, we
can have
\begin{eqnarray}
\log p(\mathbf{x} |\tilde{\bf u}_n, \Sigma_n) &=&
 \int_{\Omega_X} q(\tilde{\bf X}_n|{\bf x})
\log { p ({\bf x}, \tilde{\bf X}_n|\tilde{\bf u}_n, \Sigma_n) \over q(\tilde{\bf X}_n |{\bf x})} 
d\Omega_X~~~~~~~~
\nonumber
\\
&+& \int_{\Omega_X} D_{KL} [
q(\tilde{\bf X}_n | {\bf x}) | p(\tilde{\bf X}_n | {\bf x};\tilde{\bf u}_n, \Sigma_n)] d\Omega_X
\end{eqnarray}
where $q(\tilde{\bf X}_n|{\bf x})$ is a surrogate model, and
\begin{eqnarray}
D_{KL}[q(\tilde{\bf X}_n| {\bf x})\|{ p(\tilde{\bf X}_n |\bf x)}]
= \int_{\Omega_X}
q(\tilde{\bf X}_n|{\bf x}) \log {
q(\tilde{\bf X}_n|{\bf x}) \over
p(\tilde{\bf X}_n | {\bf x};\tilde{\bf u}_n, \Sigma_n)
}d \Omega_X
\end{eqnarray}
is the Kullback-Leibler divergence.
If we choose the surrogate model as the conditional probability,
\[
q(\tilde{\bf X}_n|{\bf x}) = p(\tilde{\bf X}_n|{\bf x};\tilde{\bf u}_n, \Sigma_n)~\to~
D_{KL} \equiv 0
\]
and thus, we have the following identity,
\begin{eqnarray}
&&\log p (\mathbf{x}|\tilde{\bf u}_n, \Sigma_n) =
\int_{\Omega_X}
p(\tilde{\bf X}_n|{\bf x};\tilde{\bf u}_n, \Sigma_n)
\log  p ({\bf x}, \tilde{\bf X}_n |\tilde{\bf u}_n, \Sigma_n) d \Omega_X
\nonumber
\\
&&
- \int_{\Omega_X}   p(\tilde{\bf X}_n|{\bf x};\tilde{\bf u}_n, \Sigma_n)
\log p(\tilde{\bf X}_n | {\bf x};\tilde{\bf u}_n, \Sigma_n) d \Omega_X
\label{eq:Identity}
\end{eqnarray}
}

Based on Eqs. (\ref{eq:G-Mixture}) and 
(\ref{eq:Identity}), we can define
the objective function $Q({\tilde{\bf u}}, \Sigma)$
for this inverse problem as
\begin{eqnarray}
&&Q({\tilde{\bf u}},\Sigma; {\tilde{\bf u}}_n,\Sigma_n)
:=
 -
 \int_{\Omega_x}\int_{\Omega_X} p\left({\tilde{\bf X}}_n |
 \mathbf{x}; {\tilde{\bf u}}_n, \Sigma_n \right)
 \nonumber
 \\
&&
\log\Bigl(
p({\tilde{\bf X}| \tilde{\bf u}_n,\Sigma_n})
p(\mathbf{x} | {\tilde{{\bf X}}_n;\tilde{\bf u}_n, \Sigma_n})
\Bigr)
d\Omega_X d\Omega_{x}
 \nonumber
 \\
 & &+ \gamma \int_{\Omega_X} \phi({\tilde{\bf u}})d \Omega_X
\nonumber
\\
 &=&
\int_{\Omega_x}\int_{\Omega_X} p\left(\mathbf{X+{\tilde{\bf u}}_n}
| \mathbf{x};{\tilde{\bf u}}_n,\Sigma_n\right)
\nonumber
\\
& \cdot & \left(
{\frac{\left\| {\bf X}+ {\tilde{\bf u}} - {\bf x} \right\| ^2}{2 \Sigma^{2}}}
+ D\log\Sigma
+ \beta W({\tilde{\bf u}})\right)
d\Omega_X d\Omega_x
\nonumber
\\
& &+ \gamma \int_{\Omega_X} \phi({\tilde{\bf u}})d \Omega_X + const.
\label{eq:entropy}
\end{eqnarray}
It can be seen from above equation that $Q$ is a functional of
the functions ${\tilde{\bf u}},~\Sigma$, in which
${\tilde{\bf u}}_n,\Sigma_n$ are known parameter values from the previous step.
Therefore, $p\left(\mathbf{X+{\tilde{\bf u}}_n} |
\mathbf{x};{\tilde{\bf u}}_n,\Sigma_n\right)$
is a fixed constant during the staggered optimization.

In FE-BL operator splitting algorithm, based on the previous step latent variables,
we first use the finite element method to calculate
the current {feasible} displacement field ${\tilde{\bf u}}_{n+1}$
that minimizes the expectation of the negative log-likelihood function,
\begin{equation}
  {\rm FE{-}Step:}~~~{\tilde{\bf u}}_{n+1} =\argmin_{{\tilde{\bf u}}
  \in \mathcal{V}} Q({\tilde{\bf u}}; {\tilde{\bf u}}_n, \Sigma_n)~,
    \label{eq:integral3}
\end{equation}
And we refer to the above solution step as the finite element step, or FE-step,
as an analog of the Maximization-step in statistics.
Once we find the new displacement field ${\tilde{\bf u}}_{n+1}$,
we can then utilize Bayes' theorem
to calculate the current step posterior probability distribution
$p\left({\tilde{\mathbf{X}}}_{n+1} | \mathbf{x}\right)$
based on the new displacement field  ${\tilde{\bf u}}_{n+1}$.
To do so, we minimize
the probabilistic elastic potential again by finding the latent variable $\Sigma_{n+1}$, i.e.
\begin{equation}
{\rm BL{-}Step:}~~\Sigma_{n+1} = \argmin_{\Sigma \in \real^{+}} Q(\Sigma ;
 {\tilde{\bf u}}_{n+1}, \Sigma_n)
\end{equation}
and we refer to this step as the Bayesian Learning step or BL-step.

It can be proven that the FE-BL algorithm can find the unique minimum of
$Q({\tilde{\bf u}},\Sigma; {\tilde{\bf u}}_n,\Sigma_n)$
in each step. The details of this prove can be found in the appendix.
\begin{theorem}
For ${\bf u} \in \mathcal{V}$ and $\Sigma \in \mathcal{R}^+$,
there exists a unique solution $({\bf u}^{\star}, \Sigma^{\star})$, such that
\[
Q ({\bf u}^{\star}, {\Sigma^{\star}}; {\tilde{\bf u}}_n, \Sigma_n) = \inf
\limits_{{\tilde{\bf u}}, \Sigma \in \mathcal{V}\oplus \mathcal{R}^+}
Q ({\tilde{\bf u}}, \Sigma ; {\tilde{\bf u}}_n, \Sigma_n) .
\]
\label{theorem: unique solution}
\end{theorem}

We note that
the posterior probability distribution $p\left({\tilde{\bf X}}_n | \mathbf{x}; {\tilde{\bf u}}_n,
\Sigma_n\right)$ can be expressed as
\begin{eqnarray}
&&p\left({\tilde{\bf X}}_n | \mathbf{x}; {\tilde{\bf u}}_n, \Sigma_n\right)
 = p\left(\mathbf{X} + {\tilde{\bf u}}_n | \mathbf{x}\right)
= \frac{p\left(\mathbf{x} | \mathbf{X}+ {\tilde{\bf u}}_n\right) p\left( \mathbf{X}+ {\tilde{\bf u}}_n\right)}
{p\left( \mathbf{x}\right)}
\nonumber
 \\
&=&
\displaystyle
{
\exp \left(
-\frac{1}{2}
\left\| \frac{{\bf X}+ {\tilde{\bf u}}_n - {\bf x}}{\Sigma_n} \right\|^2 -
\beta W({\tilde{\bf u}}_n) \right)
\over
\displaystyle
\int_{\Omega_X}
\exp \left(
-\frac{1}{2}
\left\|
\frac{ {\bf X}+ {\tilde{\bf u}}_n - {\bf x}}{\Sigma_n}
\right\|^{2} -
\beta W({\tilde{\bf u}}_n) \right) d\Omega_X
}
\end{eqnarray}
The above FE-BL algorithm
is an iterative approach which is proceeded by applying FE-step
and BL-step alternatively
until the convergence is achieved,
which is guaranteed based on the fact that the
probabilistic entropy potential is monotonously decreasing.

To show that the proposed solution of FE-BL algorithm is equivalent
to the Deformation Recovery Problem, and the FE-BL algorithm is
convergent,
we now show that the probabilistic elastic entropy potential
in a deformable continuum is a monotonously decreasing
sequence.
According to Eq. (\ref{eq:Identity}), we may first express the
Bayesian probabilistic elastic entropy potential as,
\color{black}

\begin{eqnarray}
 && {\tilde{\Pi}} ({\tilde{\bf u}}, {\Sigma};
  {\tilde{\bf u}}_n, \Sigma_n) = -
 \int_{\Omega_x}\int_{\Omega_X} p\left({\tilde{\bf X}}_n |
 \mathbf{x}; {\tilde{\bf u}}_n, \Sigma_n \right)
 \nonumber
 \\
&&
\log\Bigl(
p({\tilde{\bf X}| \tilde{\bf u}_n,\Sigma_n})
p(\mathbf{x} | {\tilde{{\bf X}}_n;\tilde{\bf u}_n, \Sigma_n})
\Bigr)
d\Omega_X d\Omega_{x}
\nonumber
\\
&+& \gamma \int_{\Omega_X} \phi({\tilde{\bf u}})d \Omega_X
\nonumber
\\
&+&
\int_{\Omega_x}\int_{\Omega_X}
    p\left({\tilde{\bf X}}_n | \mathbf{x}; {\tilde{\bf u}}_n, \Sigma_n \right)
\log p({\tilde{\mathbf{X}} | \mathbf{x}; \tilde{\bf u}_n, \Sigma_n})
d\Omega_X d\Omega_x
\nonumber
\\
&=& ~ Q({\tilde{\bf u}}, {\Sigma};{\tilde{\bf u}}_n,\Sigma_n) +
 H({\tilde{\bf u}}, {\Sigma};{\tilde{\bf u}}_n, \Sigma_n) ,
\end{eqnarray}
where the Gibbs entropy,
 $H({\tilde{\bf u}},
 {\Sigma};{\tilde{\bf u}}_n, \Sigma_n)$, is defined as
\begin{equation}
    H({\tilde{\bf u}}, {\Sigma};{\tilde{\bf u}}_n, \Sigma_n) =
        \int_{\Omega_x}\int_{\Omega_X}
        p\left({\tilde{\bf X}}_n | \mathbf{x}; {\tilde{\bf u}}_n, \Sigma_n\right)
    \log p({\tilde{\mathbf{X}} | \mathbf{x}; \tilde{\bf u}_n, \Sigma_n})
    d\Omega_X d\Omega_x.
\end{equation}

Based on the Jensen inequality, we have
\color{black}
\begin{equation}
    H({\tilde{\bf u}}_{n+1}, \Sigma_{n+1};{\tilde{\bf u}}_n, \Sigma_n)  \geq  H({\tilde{\bf u}}, \Sigma
    ;{\tilde{\bf u}}_n, \Sigma_n), ~~ ~~~ \forall ~{\tilde{\bf u}}, \Sigma.
\end{equation}
Hence we can calculate the difference between two probabilistic elastic potentials
corresponding to two adjacent deformation configurations,
\begin{eqnarray}
&&{\tilde{\Pi}} ({\tilde{\bf u}}_{n+1}, \Sigma_{n+1}) - {\tilde{\Pi}} ({\tilde{\bf u}}_n, \Sigma_{n})
\nonumber
\\
&=&
Q({\tilde{\bf u}}_{n+1},\Sigma_{n+1} {; {\tilde{\bf u}}_n, \Sigma_n}) - Q({\tilde{\bf u}}_n,\Sigma_n {; \tilde{\bf u}_n, \Sigma_n}) \\
&&+H({\tilde{\bf u}}_{n+1}, \Sigma_{n+1} {; \tilde{\bf u}_n, \Sigma_n}) -
 H({\tilde{\bf u}}_n, \Sigma_n {; \tilde{\bf u}_n, \Sigma_n})
 \nonumber
 \\
&\leq& Q({\tilde{\bf u}}_{n+1},\Sigma_{n+1} {; \tilde{\bf u}_n, \Sigma_n}) - Q({\tilde{\bf u}}_n,\Sigma_n {; \tilde{\bf u}_n, \Sigma_n})  \leq 0.
    \label{eq:Pi-E}
\end{eqnarray}
Hence during the BL-FE iteration, the probabilistic elastic potential is monotonously decreasing.

After showing the monotonous decreasing property of the probabilistic elastic potential,
we focus on the staggered BL-step/FE-step of the VBI-FEM computational algorithm,
which may be formulated as the following optimization problem.\\

\section{Variational Bayesian learning finite element formulation and implementation}
\label{sec: VE-FEM formulation and implementation}

We now present the FE-step formulation and implementation of VBI-FEM.
It is noted that the underline elastic continuum itself
is not necessarily statistical or stochastic.
It maybe just the epistemic uncertainty making
the inverse VBI-FEM formulation probabilistic.

Assume that in an elastic continuum, the stress-strain state is
$({ \sigma} ({\bf x}), { \epsilon} ({\bf x})),
\forall {\bf x} \in \Omega$.
Based on Theorem \ref{theorem: unique solution},
to maximize the Gibbs entropy,
we first consider the following variational formulation at n-th FE step,
\begin{eqnarray}
&&\delta_{{\tilde{\bf u}}}
 Q ({\tilde{\bf u}}_{n+1}; {\tilde{\bf u}}_{n}, \Sigma_n)
\nonumber
\\
&=& \beta \int_{\Omega_X}\int_{\Omega_x}
p\left(\mathbf{X} +{\tilde{\bf u}}_n | \mathbf{x}\right) \cdot
{ \sigma}_{n+1} ({\tilde{\bf X}}):\delta { \varepsilon}_{n+1}
d\Omega_X d\Omega_x
\nonumber
\\
&+&
\frac{1}{\Sigma^2_n}\int_{\Omega_X}\int_{\Omega_x}
p\left(\mathbf{X}+{\tilde{\bf u}}_n | \mathbf{x}\right)\cdot
(\mathbf{X}+ {\tilde{\bf u}}_n-\mathbf{x})\delta
{\tilde{\bf u}}_{n+1}
d\Omega_X d\Omega_x
\nonumber
\\
&+& \gamma \int_{\Omega_X}  {\tilde{\bf u}}_{n+1} \cdot \delta {\tilde{\bf u}}_{n+1}
    d \Omega_X =0 .
\label{eq:weak form2b}
\end{eqnarray}
The probabilistic Cauchy stress $\bar{ \sigma}_{n+1} ({\tilde{\bf X}})$
at the n-th iteration is defined as,
\begin{equation}
\bar{ \sigma}_{n+1} ({\tilde{\bf X}})=
\int_{\Omega_x}
 p\left(\mathbf{X}+{\tilde{\bf u}}_n | \mathbf{x}\right) \cdot
    { \sigma}({\bf X}+{\tilde{\bf u}}_{n+1}) d \Omega_x ,
\end{equation}
and the probabilistic body
force $\bar{{\bf b}}_n({\tilde{\bf X}})$ at the n-th iteration is defined as
\begin{equation}
\bar{\bf b}_n ({\tilde{\bf X}}) =       \frac{1}{\Sigma^2}\int_{\Omega_x}
    p\left(\mathbf{X}+{\tilde{\bf u}}_n | \mathbf{x}\right)\cdot
    ({\bf x} - \mathbf{X}-{\tilde{\bf u}}_{n})
   d\Omega_x .
\end{equation}

Furthermore, to simplify Eq. (\ref{eq:weak form2b}),
we define an equivalent probability $\bar{p}({\tilde{\bf X}})$ at the n-th iteration as
\begin{equation}
    \bar{p}_n({\tilde{\bf X}}) = \int_{\Omega_x}
 p\left(\mathbf{X}+{\tilde{\bf u}}_n | \mathbf{x}
 \right)
     d \Omega_x .
\end{equation}
Hence, Eq. (\ref{eq:weak form2b}) can be rewritten as follows,
\begin{eqnarray}
&&\beta\int_{\Omega_X} \bar{{ \sigma}}_{n+1}: \delta {\bfg \epsilon}_{n+1} d \Omega_X
+ \gamma \int_{\Omega_X} {\tilde{\bf u}}_{n+1} :\delta {\tilde{\bf u}}_{n+1} d \Omega_X
~~~~~~~~
\nonumber
\\
&&+ \frac{1}{\Sigma^2_n} \int_{\Omega_X} \bar{p}_n {\tilde{\bf u}}_{n+1} :\delta {\tilde{\bf u}}_{n+1} d \Omega_X
- \int_{\Omega_X} \bar{\bf b}_n :\delta {\tilde{\bf u}}_{n+1} d \Omega_X = {\bf 0},
\nonumber
\\
\end{eqnarray}
which is the Galerkin weak formulation or the virtual work formulation for the VBI-FEM.
Considering finite element interpolation e.g. \cite{hughes2012finite},
\begin{eqnarray}
&&{\tilde{\bf u}}^h_{n+1} = [{\bf N}][{\tilde{\bf u}}_{n+1}],
~~\delta {\tilde{\bf u}}^h_{n+1} = [{\bf N}] [ \delta {\tilde{\bf u}}_{n+1}],~~{\rm and}~~
\\
\nonumber
\\
&&{ \epsilon}^h_{n+1} =[{\bf B}] [{\bf u}_{n+1}]
\end{eqnarray}
the probabilistic Galerkin weak formulation can be discretized as follows,
\begin{eqnarray}
&&\int_{\Omega_X}  [{\bf B}]^T \bar{\sigma}_{n+1}  d \Omega_X
+ \gamma \int_{\Omega_X}  [{\bf N}] {\tilde{\bf u}}_{n+1} d \Omega_X~~~~~~~~~
\nonumber
\\
\nonumber
\\
&&+ \frac{1}{\Sigma^2_n} \int_{\Omega_X}  [{\bf N}] \bar{p}_n {\tilde{\bf u}}_{n+1}
d \Omega_X
- \int_{\Omega_X} [{\bf N}]\bar{\bf b}_n  d \Omega_X
= 0 ,
\label{eq:weak form}
\end{eqnarray}
 where $[{\bf N}]$ is the global shape function matrix, $[{\bf B}]$ is the global
 strain-displacement matrix, and $[{\tilde{\bf u}}_n]$
 is the finite element nodal displacement
 at the n-th iteration.

Then the probabilistic stress $\bar{ \sigma}_n$ can be defined as
\begin{equation}
\bar{{ \sigma}}^h_{n+1}({\bf X}) = \int_{\Omega_X}
\bar{p}_n({\bf X}) [{\bf D}][{\bf B}] [{\bf d}_{n+1}] d \Omega_X,
\end{equation}
where $[{\bf D}]$ is the the elastic modulus matrix and $[{\tilde{\bf u}}]$
is the global nodal displacement vector.
Let the internal force vector ${\bf f}^{int}_n$ and the external force vector ${\bf f}^{ext}_n$
at the n-th iteration be expressed as
\begin{eqnarray}
{\bf f}^{int}_{n+1} &=&
\int_{\Omega_X}  [{\bf B}]^T {\bar{\sigma}}_{n+1}  d \Omega_X
+ \gamma \int_{\Omega_X}  [{\bf N}] {\tilde{\bf u}}_{n+1} d \Omega_X
\nonumber
\\
&&+ \frac{1}{\Sigma^2_n} \int_{\Omega_X}  [{\bf N}] \bar{p}_n {\tilde{\bf u}}_{n+1}
d \Omega_X
\end{eqnarray}
and
\begin{equation}
{\bf f}^{ext}_{n+1} = \int_{\Omega_X} [{\bf N}] \bar{{\bf b}}_n  d \Omega_X.
\end{equation}
Then, Eq. (\ref{eq:weak form}) can also be expressed as,
\begin{equation}
{\bf f}^{int}_{n+1}-{\bf f}^{ext}_{n+1}=0,
\label{eq:balance of linear momentum}
\end{equation}
which is the discrete version of the equilibrium equation in continuum mechanics.

For each element in the reference configuration, the element nodal force is defined as
\begin{eqnarray}
{\bf f}^{int}_e &=&
\int_{\Omega_{e}}  [{\bf B}]^T \bar{{ \sigma}}  d \Omega_{e}
+ \gamma \int_{\Omega_{e}}  [{\bf N}] {\tilde{\bf u}} d \Omega_{e}
+ \frac{1}{\Sigma^2} \int_{\Omega_e}  [{\bf N}] \bar{p} {\tilde{\bf u}}
d \Omega_e
\nonumber
\\
&=&
\beta[\bar{\bf K}_e][{\tilde{\bf u}}_e] + \gamma [ {\bf M}_e][{\tilde{\bf u}}_e]
+ \frac{1}{\Sigma^2} [\bar{\bf M}_e][{\tilde{\bf u}}_e],
\end{eqnarray}
where $[{\bf d}]_e$ is element nodal displacement vector,
and  $[\bar{{\bf K}_e}]$, $[{\bf M}_e]$ and $[\bar{\bf M}_{e}]$
are defined as
\begin{equation}
[\bar{\bf K}_{e}]=\int_{\Omega_{e}}
\bar{p}({\bf X})
[{\bf B}]^{T}[{\bf D}][{\bf B}]  d \Omega_{e},
\end{equation}
\begin{eqnarray}
[{\bf M}_{e}]=\int_{\Omega_{e}}
[{\bf N}]^T[{\bf N}]
d \Omega_{e},~~{\rm and}~~
[\bar{\bf M}_{e}]=\int_{\Omega_{e}}
\bar{p}({\bf X})[{\bf N}]^T[{\bf N}]
d \Omega_{e}.
\end{eqnarray}
Correspondingly, the equivalent element nodal force $[{\bf b}_e]$ can be defined
as follows,
\begin{equation}
[{\bf b}_e] = [{\bf f}^{ext}_e] = \int_{\Omega_e} [{\bf N}] \bar{{\bf b}}({\bf X})  d \Omega_e ~.
\end{equation}

After obtaining the element equivalent stiffness matrix $[\bar{\bf K}_e]$,
the global equivalent stiffness matrix $[\bar{\bf K}]$ can be assembled
by using the connectivity matrix $[{\bf L}_e]$,
\begin{equation}
    [\bar{\bf K}] = \ass [{\bf L}_e]^T [\bar{\bf K}_e] ,
    \label{eq:global K}
\end{equation}
where $N_e$ is the total number of elements, and $\ass$ element assembly operator
\cite{hughes2012finite}.
Similarly, the global matrices
$[{\bf M}], ~[\bar{\bf M}]$ and $[{\bf b}]$ can also be assembled as

\begin{equation}
[{\bf M}] = \ass [{\bf L}_e]^T [{\bf M}_e],
\end{equation}
\begin{equation}
[\bar{{\bf M}}] = \ass [{\bf L}_e]^T [\bar{{\bf M}_e}],~~{\rm and}~~
\end{equation}
\begin{equation}
[{\bf b}] = \ass [{\bf L}_e]^T [[{\bf b}_e].
    \label{eq:global matrices}
\end{equation}

Therefore, substituting Eq. (\ref{eq:global K}) and Eq. (\ref{eq:global matrices})
into Eq. (\ref{eq:weak form}), one can obtain the equilibrium equation at the n-th iteration,
\begin{equation}
[{\bf f}^{int}_{n+1}] - [{\bf f}^{ext}_{n+1}] = \beta[\bar{{\bf K}}_{n}][{\tilde{\bf u}}_{n+1}]
+ \gamma [{\bf M}][{\tilde{\bf u}}_{n+1}]
+ \frac{1}{\Sigma^2_n} [\bar{{\bf M}}_n][{\tilde{\bf u}}_{n+1}] - [{\bf b}_n] =0.
\label{eq:fem1}
\end{equation}
Let $[\tilde{\bf K}_n]$ be the total stiffness matrix at the n-th iteration, i.e.
\begin{equation}
    [\tilde{\bf K}_n] =
    \beta[\bar{{\bf K}}_n] + \gamma [{\bf M}]
+ \frac{1}{\Sigma^2_n} [\bar{\bf M}_n],
\label{eq:total stiffness matrix}
\end{equation}
by substituting Eq. (\ref{eq:total stiffness matrix}) into Eq. (\ref{eq:fem1}),
the discrete FEM governing equation at the n-th iteration becomes,
\begin{equation}
     [\tilde{\bf K}_n][{\tilde{\bf u}}_{n+1}] = [{\bf b}_n],
\end{equation}
which has the same form as that in conventional static finite element analysis.
Finally, the nodal displacement vector $[{\tilde{\bf u}}]$
can be solved by the same methods used in classical finite element, such as
\begin{equation}
    [{\tilde{\bf u}}_{n+1}] =  [\tilde{\bf K}_n]^{-1}[{\bf b}_n].
    \label{eq: disp u}
\end{equation}

After ${\tilde{\bf u}}_{n+1}$ is found,
we consider the variational formulation for the statistical parameter at the (n+1)-th iteration:
\begin{eqnarray}
&&\delta_{\Sigma} Q ({\tilde{\bf u}}_{n+1},\Sigma_{n+1}) =
    \int_{\Omega_x}\int_{\Omega_X} p\left(\mathbf{X} +{\tilde{\bf u}}_{n+1}| \mathbf{x}\right)
    ~~~~~~~~~~~~~~~~~
    \nonumber
    \\
    &&\cdot \left(
 -{\frac{\left\| {\bf X}+ {\tilde{\bf u}}_{n+1} - {\bf x} \right\| ^2}{\Sigma_{n+1}^{3}}}+
    \frac{D}{\Sigma_{n+1}}\right)
    \nonumber
\cdot \delta \Sigma_{n+1}
    d\Omega_X d\Omega_x = 0 ,
    \nonumber
    \\
    &&~ \forall \delta\Sigma \in \mathbb{R}^+
    \label{eq: sigma equation1}
\end{eqnarray}
Note that the above equation is valid for any arbitrary $\delta\Sigma \in \mathbb{R}^+$.
As a result, the Eq. (\ref{eq: sigma equation1}) implies that
\begin{eqnarray}
  &&  \int_{\Omega_x}\int_{\Omega_X} p\left(\mathbf{X}+{\tilde{\bf u}}_{n+1} | \mathbf{x}\right) \cdot
  \nonumber
  \\
  &&
  \left(
 -{\frac{\left\| {\bf X}+ {\tilde{\bf u}}_{n+1} - {{\bf x}} \right\| ^2}{\Sigma_{n+1}^{3}}}+
    \frac{D}{\Sigma_{n+1}}\right)\cdot
    d\Omega_X d\Omega_x = 0 .
    \label{eq: sigma equation2}
\end{eqnarray}

\begin{figure}[ht]
\begin{center}
\includegraphics[width=\columnwidth]{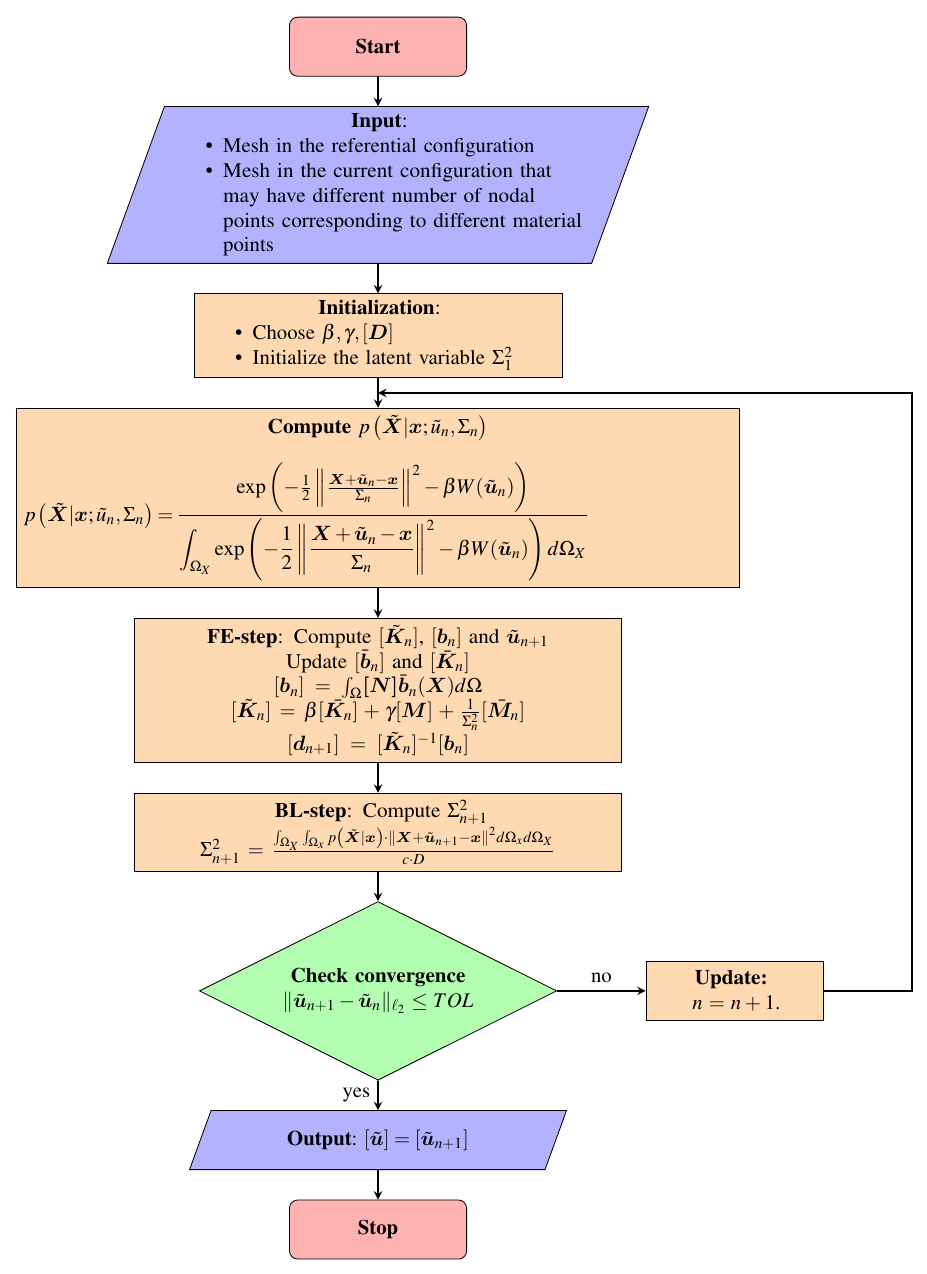}
\end{center}
\caption{{Flowchart for VBI-FEM inverse recovery algorithm for continuum deformation mapping.}}
\label{fig:procedure}
\end{figure}

After some algebraic manipulations, Eq. (\ref{eq: sigma equation2}) can be solved as
\begin{eqnarray}
    \Sigma^2_{n+1} &=& \frac{\int_{\Omega_X}\int_{\Omega_x} p\left(\mathbf{X}
    +{\tilde{\bf u}}_n| \mathbf{x}\right)\cdot
    \left\| {\bf X}+ {\tilde{\bf u}}_{n+1} - {\bf x} \right\| ^2d\Omega_x d\Omega_X}
    {D \int_{\Omega_X}\int_{\Omega_x}p\left(\mathbf{X}+{\tilde{\bf u}}_{n+1}
    | \mathbf{x}\right)d\Omega_x d\Omega_X}
    \nonumber
    \nonumber
    \\
    &=& \frac{\int_{\Omega_X}\int_{\Omega_x} p\left(\mathbf{X}+{\tilde{\bf u}}_{n+1} | \mathbf{x}\right)\cdot
    \left\| {\bf X}+ {\tilde{\bf u}}_{n+1} - {\bf x} \right\|^2 d\Omega_x d\Omega_X}
    {c \cdot D} ,
    \nonumber
    \\
\label{eq: updated sigma}
\end{eqnarray}
where constant $c$ represents the area of the mesh in the current configuration.
After solving the nodal displacement vector $[{\tilde{\bf u}}_{n+1}]$ in Eq. (\ref{eq: disp u}),
we shall test the convergence criterion, i.e. $\| {\tilde{\bf u}}_{n+1} - {\tilde{\bf u}}_n \|_{\ell_2} \le TOL$,
where $TOL$ is a preset tolerance number. If the convergence criterion is met,
We obtain the inverse solution of the displacement field: ${\tilde{\bf u}} = {\tilde{\bf u}}_{n+1}$.
Otherwise, we let $n=n+1$ and proceed to the iteration $n+1$, until it convergence.
If the two point sets have a one-to-one correspondence, one can use the convergence criterion,
$\| \tilde{\bf X}_{n} - {\bf x} \|_{\ell_2} \le TOL$, which is easy to be satisfied in computations.
In summary, the flowchart of the VBI-FEM computational algorithm is shown in Fig. \ref{fig:procedure}.

In fact, based on Eq.(\ref{eq:Identity}, we can write
\begin{eqnarray}
- p({\bf x}|\tilde{\bf u}_n, \Sigma_n)
= \int_{\Omega_X} p(\tilde{\bf X}_n |{\bf x};\tilde{\bf u}_n, \Sigma_n)
\log \Bigl(
{
p(\tilde{\bf X}_n |{\bf x};\tilde{\bf u}_n, \Sigma_n)
\over
p(\tilde{\bf X}_n | {\bf x};\tilde{\bf u}_n, \Sigma_n)
}
\Bigr) d \Omega_X
\end{eqnarray}
as a form of the Kullback-Leibler divergence or a relative entropy.
This is the variational inference origin of the maximum principle stated in Theorem \ref{theorem: unique solution}.

\section{Numerical examples and applications}
\label{sec:example}
In this Section, we present
two VBI-FEM numerical inverse solutions to recover the deformation mappings.
The first example concerns the inverse problem of a two-dimensional thin plate with a hole problem.
The material of the plate is made of linear elastic material.
The second example is a three-dimensional deformation recovery
problem of a solid cylinder, which is
made of an inelastic material, and the forward problem
concerns with plasticity deformation and fracture propagation.

\subsection{A thin plate with a hole}
\label{sec: example plate with a hole}

\begin{figure}[ht]
\centering
\begin{subfigure}{.45\columnwidth}
  \centering
  \includegraphics[width=0.75\columnwidth]{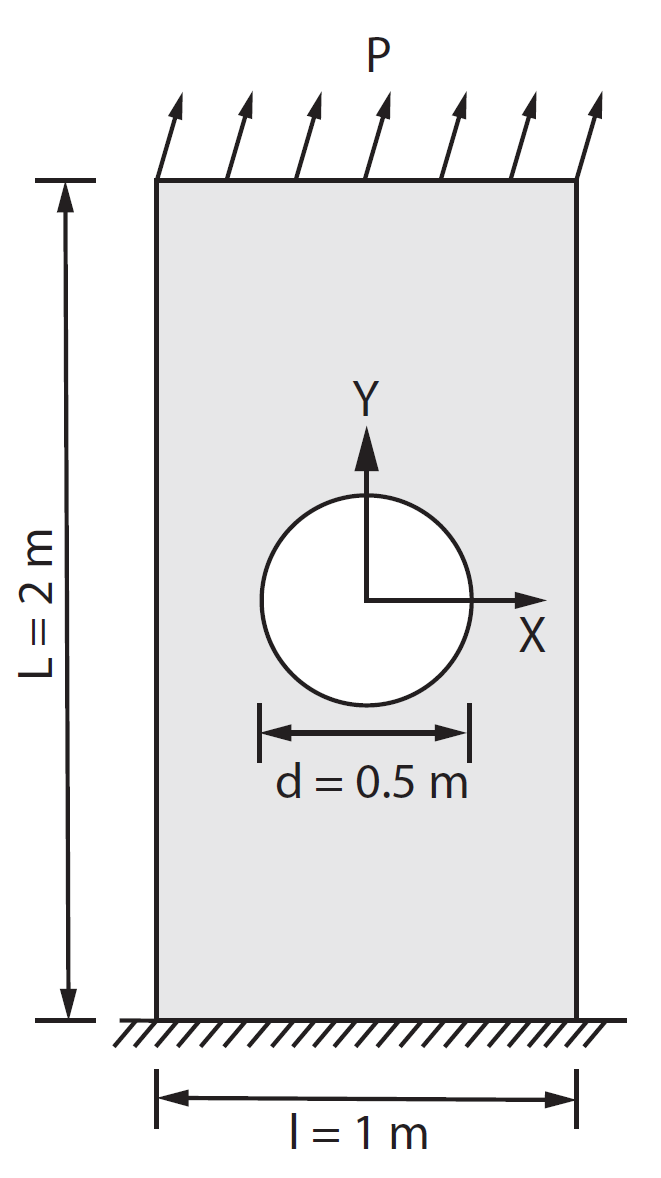}
  \caption{}
\end{subfigure}
\begin{subfigure}{.45\columnwidth}
  \centering
  \includegraphics[width=\columnwidth]{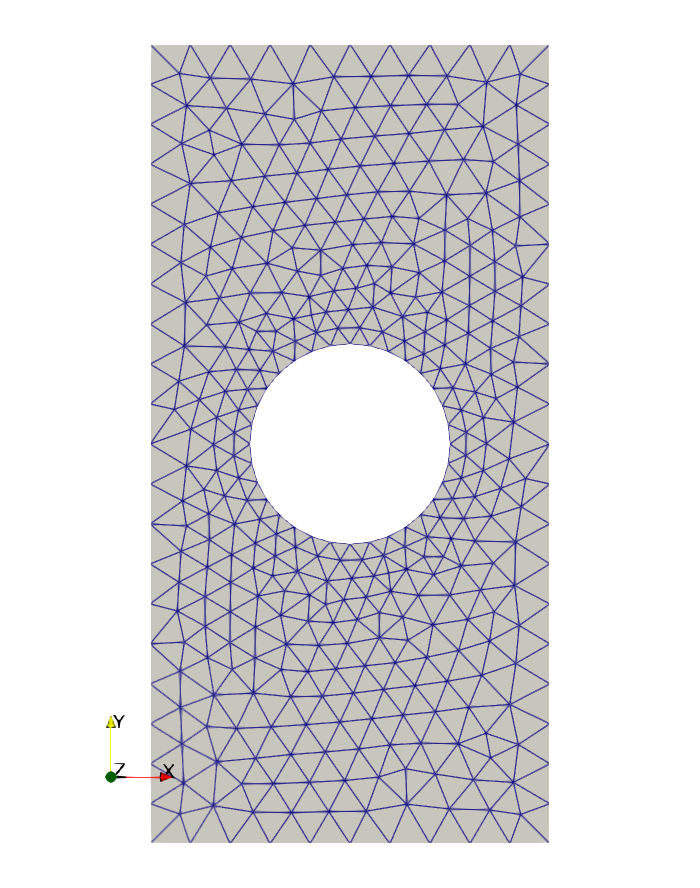}
  \caption{}
\end{subfigure}
\begin{subfigure}{.45\columnwidth}
  \centering
  \includegraphics[width=\columnwidth]{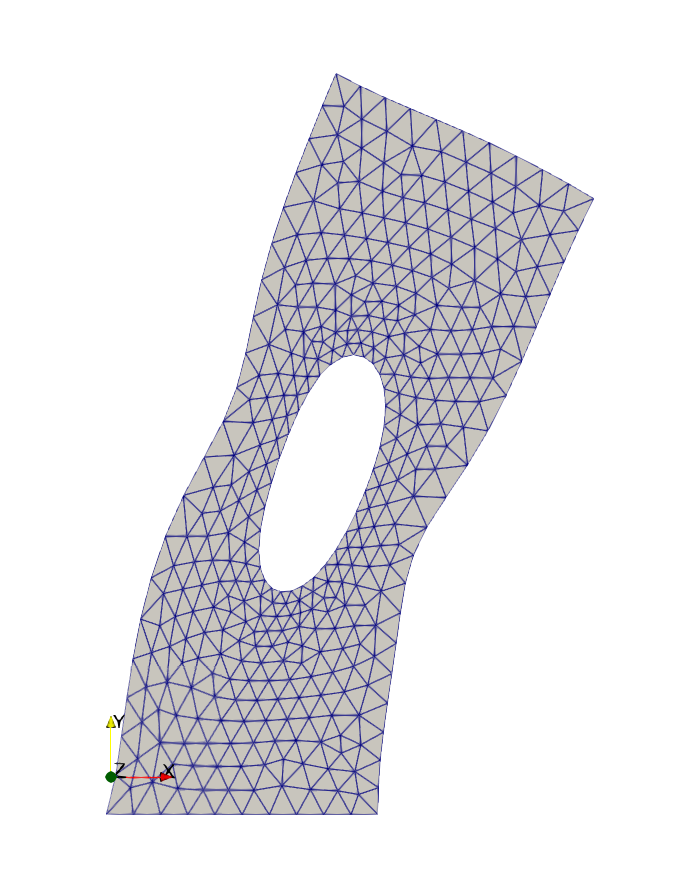}
  \caption{}
\end{subfigure}
\begin{subfigure}{.45\columnwidth}
  \centering
  \includegraphics[width=\columnwidth]{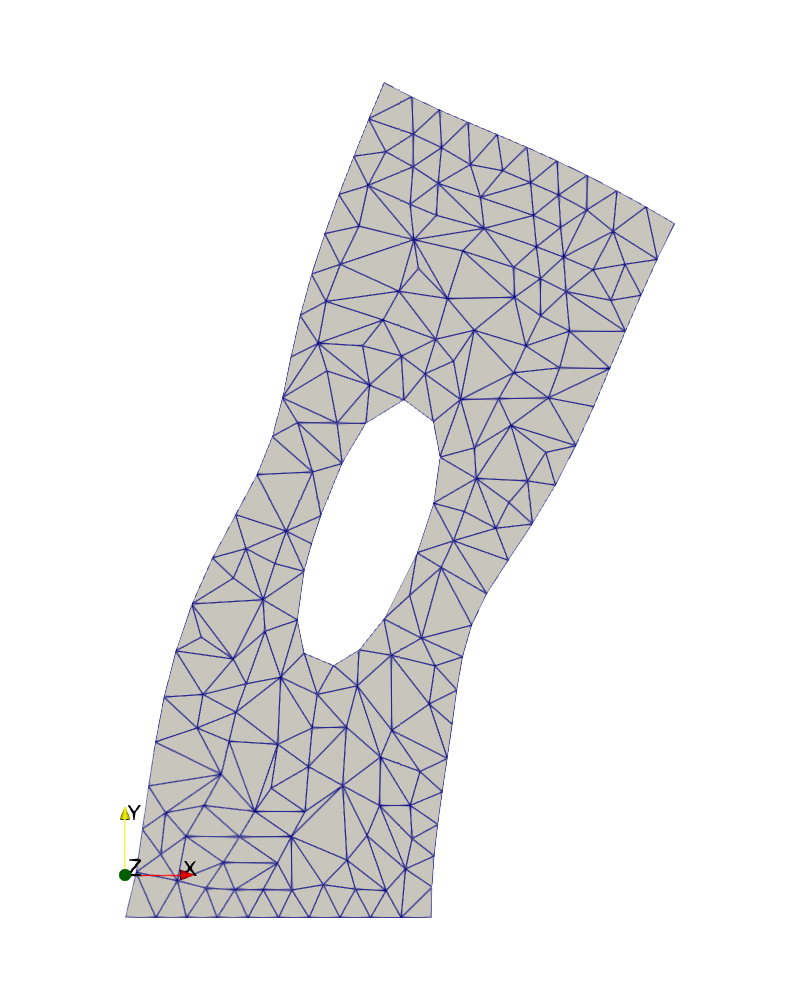}
  \caption{}
\end{subfigure}
\caption{ The finite element meshes for the thin plate example:
(a)The setup of the thin plate with a hole example;
(b) Mesh in the referential (undeformed) configuration;
(c) Mesh in the final (deformed) configuration (fine);
(d)  Mesh in the final (deformed) configuration (coarse).}
\label{fig:Plate with hole meshes}
\end{figure}

To verify the effectiveness of the proposed VBI-FEM, a thin plate with a hole under
oblique traction is conducted.
The dimensions of the thin plate are shown in Fig. \ref{fig:Plate with hole meshes}(a),
where $L = 2  ~m$, $l = 1 ~m$, and the thickness of the plate is $1~cm$.
A circular hole with a diameter of $d = 0.5 ~m$ is positioned in the center of the thin plate.

This thin plate is fixed at the bottom and an oblique uniformly distributed load ${\bf p}$
is applied at the top of the plate which reads:
$    p_x = 0.01~MPa,~~~ p_y = 0.1~MPa $
where $p_x$ and $p_y$ are the $x$ and $y$ components of the distributed
load ${\bf p}$ respectively.
The constitutive model of this plate is chosen as isotropic linear elasticity
with the Lam\'e constants $\lambda = 0.58~MPa$ and $\mu = 0.38~MPa$.

Then the initial configuration of the thin plate is discretized into
712 linear triangular elements by using Gmsh \cite{geuzaine2009gmsh}
as shown in Fig.  \ref{fig:Plate with hole meshes}(b).
This mesh is generated uniformly on the boundary with mesh size equals $0.2 ~m$ and local
refinement around the circular hole is performed to capture the high stress and strain
(see Fig. \ref{fig:Plate with hole meshes}(b)). Finally, a finite element analysis
is conducted by using the open-source FEM package MFEM \cite{mfem}.
The data points in the final configuration are generated in terms
of FEM meshes as well, which are shown in Fig. \ref{fig:Plate with hole meshes}(c) (fine)
and Fig.  \ref{fig:Plate with hole meshes}(d) (coarse).

The proposed VBI-FEM is then used to recover the inverse deformation mapping.
The finite element nodal point data of the initial and final configurations, as illustrated
in Fig. \ref{fig:Plate with hole meshes}, are given as the inputs to the VBI-FEM.
The constitutive model of VBI-FEM is also chosen as isotropic linear elasticity.
The mesh nodal points in the initial undeformed
configuration are chosen as the Gaussian Mixture Model
centroid points,
and the mesh nodal points in the deformed configuration are set as the reference data points.
Therefore, the mesh in the initial configuration moves towards the mesh in the deformed
configuration to recover the inverse deformation.
This is the forward solution of the inverse problem.
Since there are two sets of data of mesh nodal points in the deformed configurations, the fine and coarse,
there will be two numerical solutions for the inverse problem.
To distinguish the two, we label them as case I (fine) and case II (coarse).

In order to better illustrate the capacity of VBI-FEM,
two different cases are conducted with increasing complexity.
In the first case, the same Lam\'e constants,  $\lambda = 0.58~MPa$
and $\mu = 0.38~MPa$, and the same mesh utilized
in FEA are given as the inputs to the VBI-FEM model.
For the second case, the Lam\'e constants are chosen as  $\lambda = 0.001~MPa$ and $\mu = 0.001~MPa$,
and the mesh in the final configuration is regenerated to a coarse one with only 317 linear triangular elements
 (see Fig. \ref{fig:Plate with hole meshes} (d)).
The hyperparameters for the two cases are: $\beta: \num{6e-7}$ and $\num{8e-4}$;
$\gamma: \num{1e-7} $ and $\num{1e-5}$, and the maximum number of iterations
are $200$ and $450$ respectively.

\begin{figure}[ht]
\centering
\begin{subfigure}{.45\columnwidth}
  \centering
  \includegraphics[width=\columnwidth]{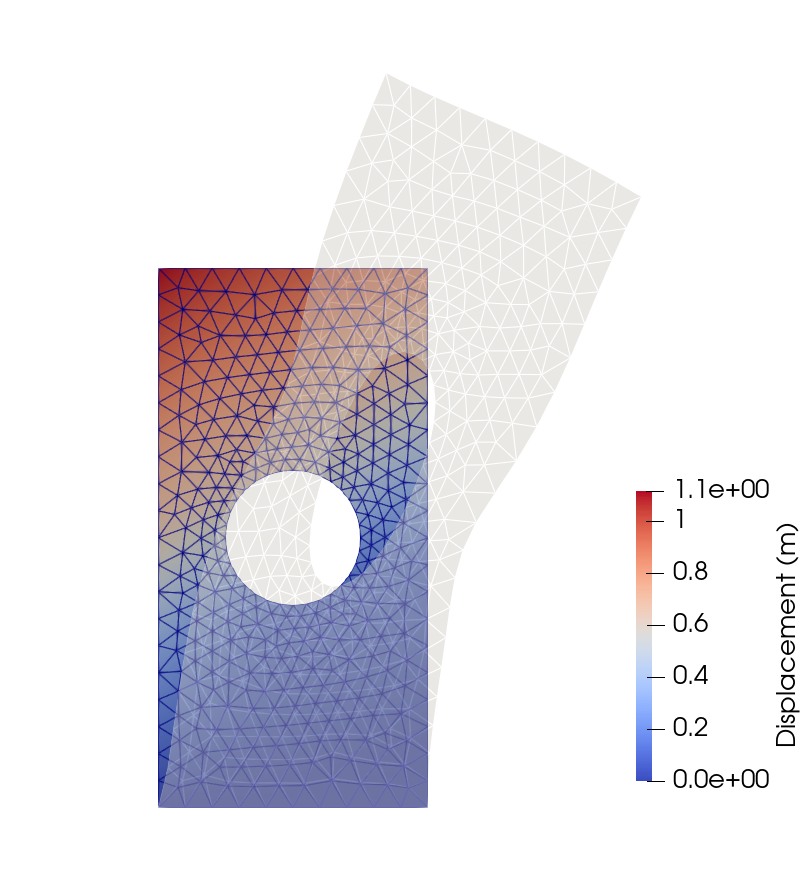}
  \caption{Initialization}
\end{subfigure}
\begin{subfigure}{.45\columnwidth}
  \centering
  \includegraphics[width=\columnwidth]{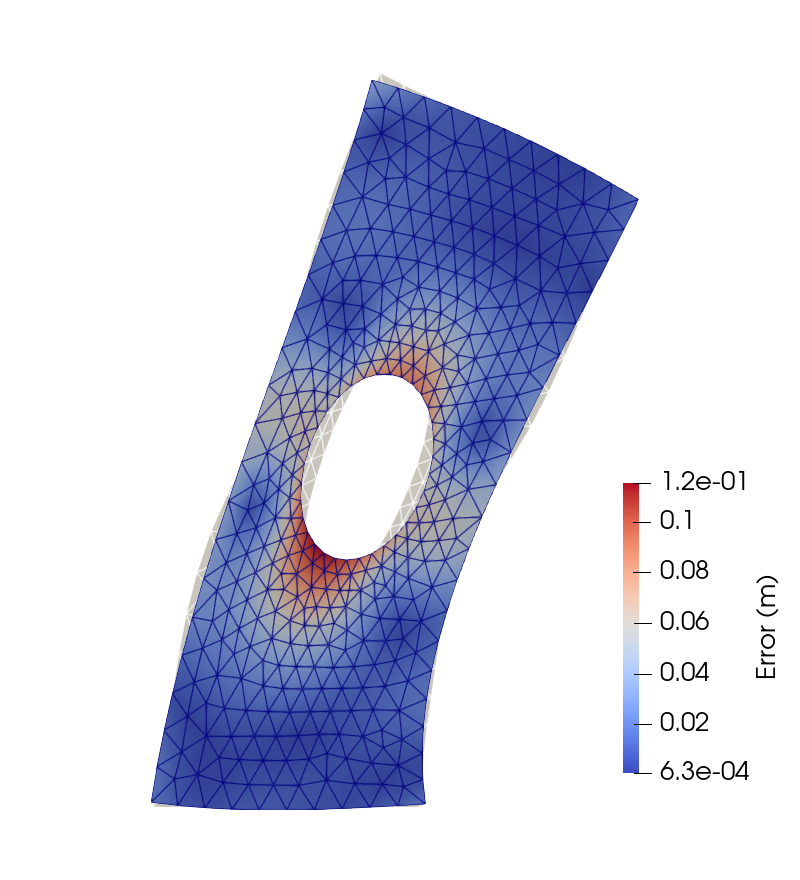}
  \caption{CPD result}
\end{subfigure}
\begin{subfigure}{.45\columnwidth}
  \centering
  \includegraphics[width=\columnwidth]{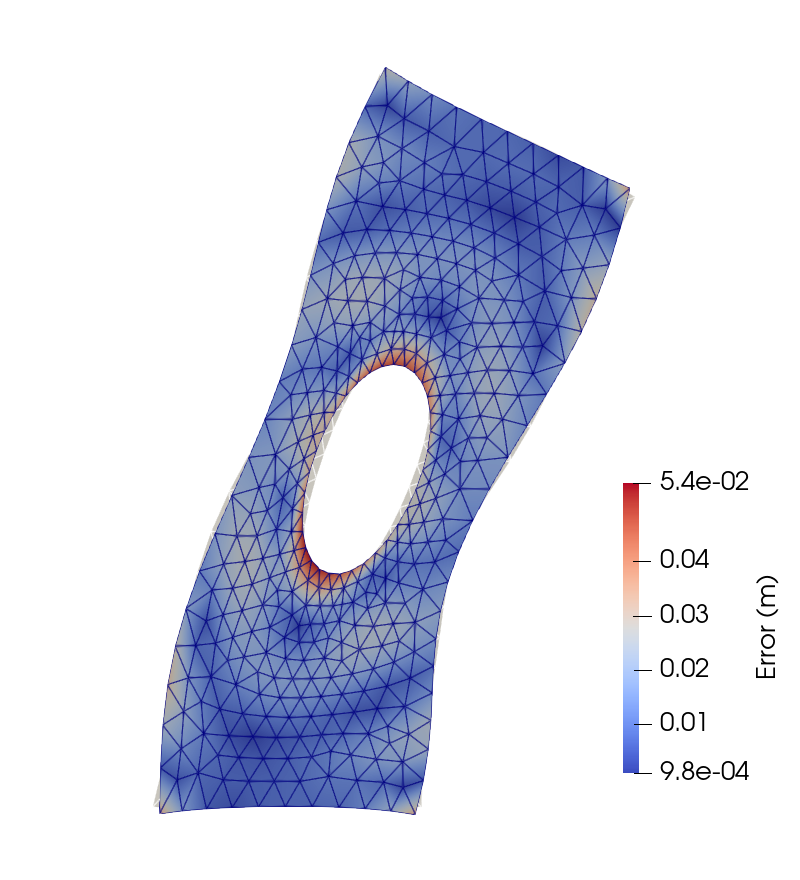}
  \caption{PR-GLS result}
\end{subfigure}
\begin{subfigure}{.45\columnwidth}
  \centering
  \includegraphics[width=\columnwidth]{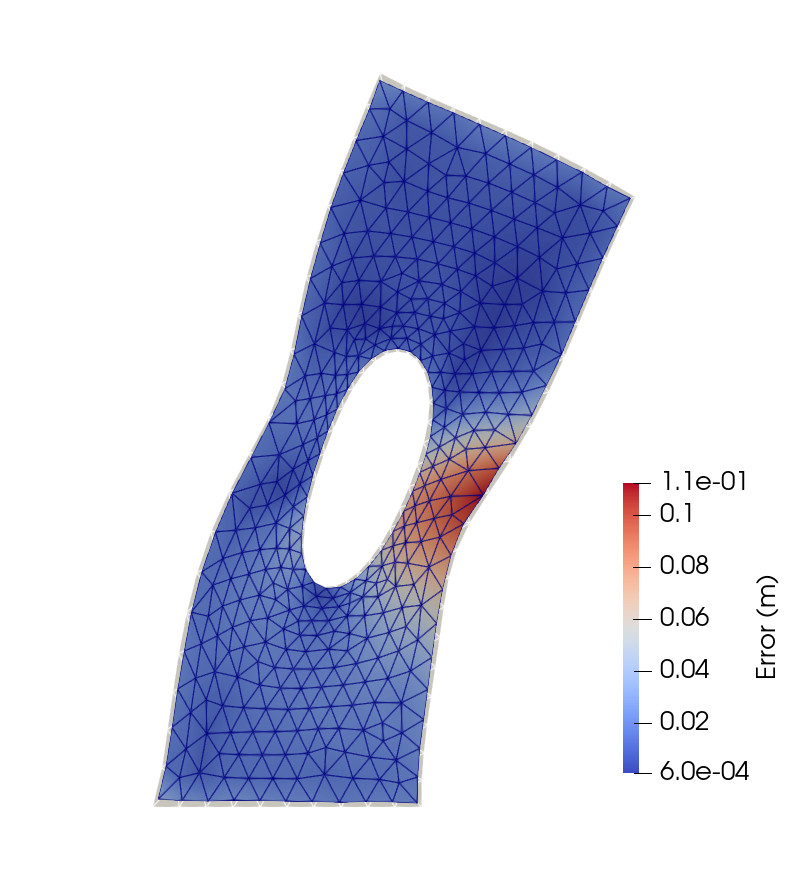}
  \caption{VBI-FEM result}
\end{subfigure}
\caption{The inverse recovery results of the deformation mapping ${ \psi} ({\bf u} ({\bf X}))$
for a plate with a hole in case I.}
\label{fig:Example 1 disp recovery process}
\end{figure}

Figure \ref{fig:Example 1 disp recovery process} presents the displacement recovery result for case I.
The final configuration is shown in gray as the true underlying deformation mapping.
The initial configuration which is color-coded by the magnitude of error along the matching process.
In order to better exemplify the accuracy of deformation recovery,
the average nodal deformation error $\bar{e}$ is evaluated, which is given as
\begin{equation}
    \bar{e} = \frac{1}{n} \sum_{i = 1}^{n} e_i
    =\frac{1}{n} \sum_{i = 1}^{n} ||{\bf x}_i^{FEA} - {\bf x}_i^{recovery}||_{2}
    \label{eq:error-calculation}
\end{equation}
where $e_i$ is the nodal displacement error for the $i$th node,
$n$ is the number of nodes in the given mesh, ${\bf x}_i^{FEA}$
is the final position for $i$th node which is calculated by FEA, and
${\bf x}_i^{recovery}$ is the recovered final position for $i$th node by
the VBI-FEM model. In Eq. (\ref{eq:error-calculation}),
 $||\cdot||_2$ represents the Euclidean norm.

\begin{figure}[ht]
\centering
\begin{subfigure}{.45\columnwidth}
  \centering
  \includegraphics[width=\columnwidth]{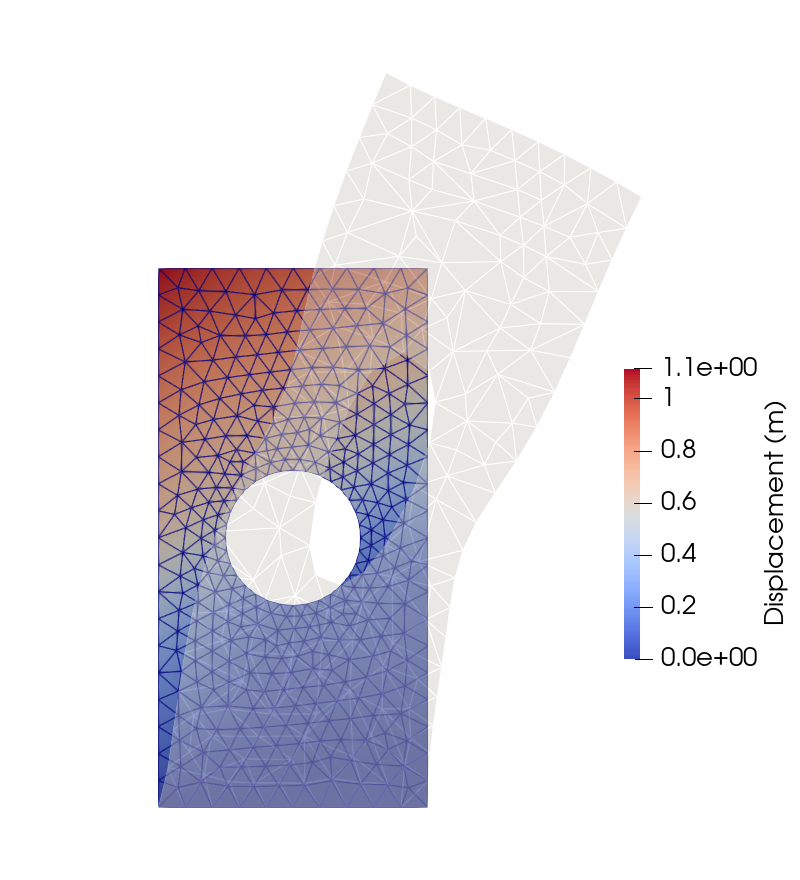}
  \caption{Initialization}
\end{subfigure}
\begin{subfigure}{.45\columnwidth}
  \centering
  \includegraphics[width=\columnwidth]{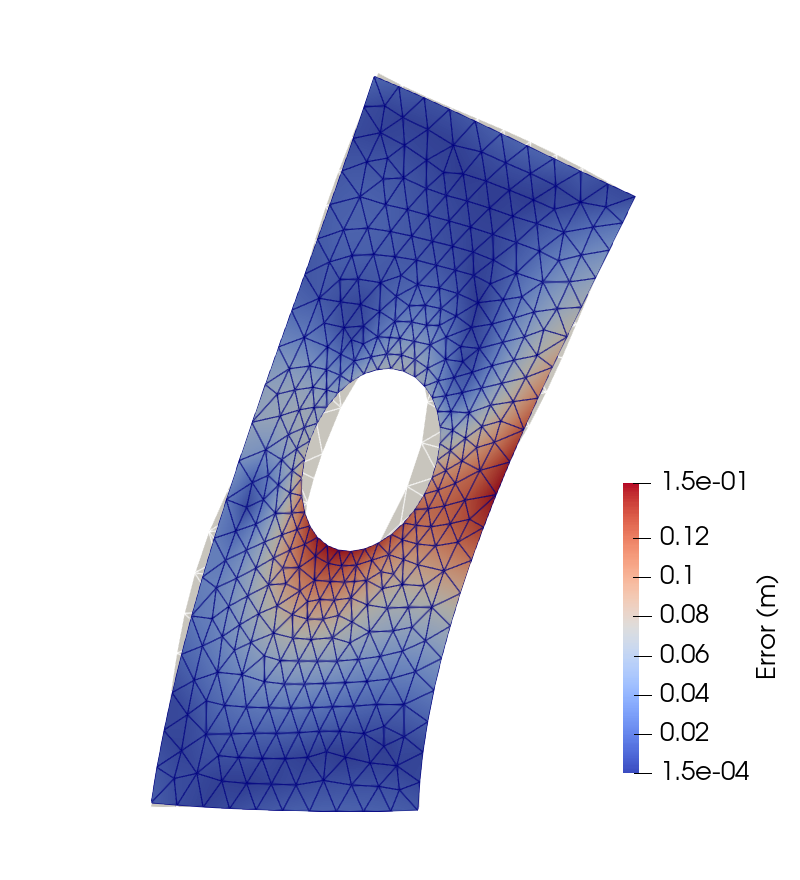}
  \caption{CPD result}
\end{subfigure}
\begin{subfigure}{.45\columnwidth}
  \centering
  \includegraphics[width=\columnwidth]{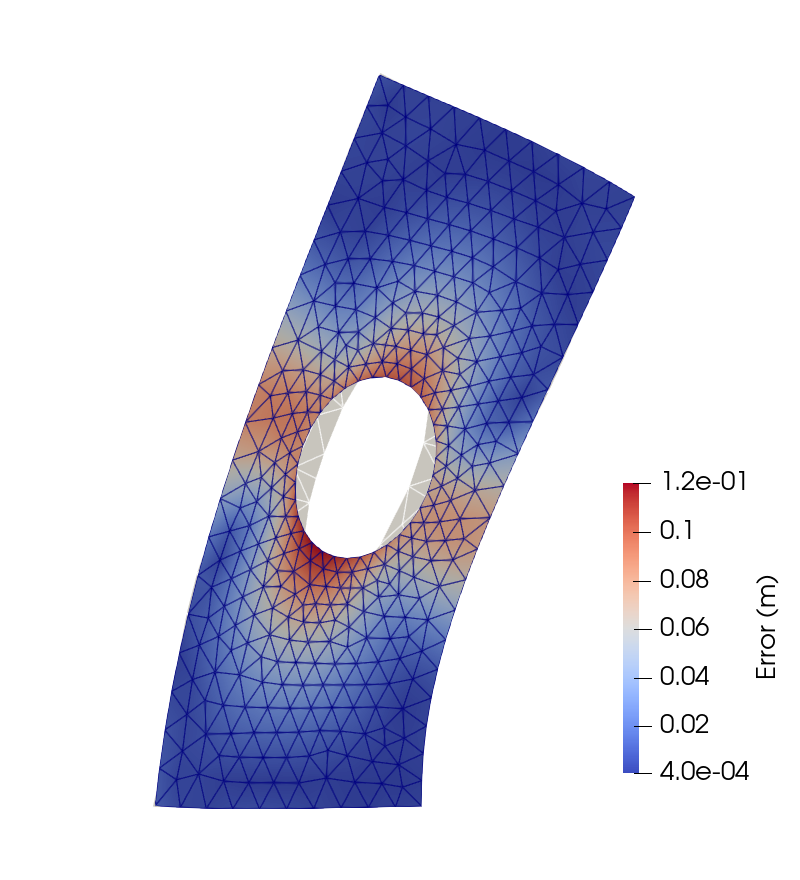}
  \caption{PR-GLS result}
\end{subfigure}
\begin{subfigure}{.45\columnwidth}
  \centering
  \includegraphics[width=\columnwidth]{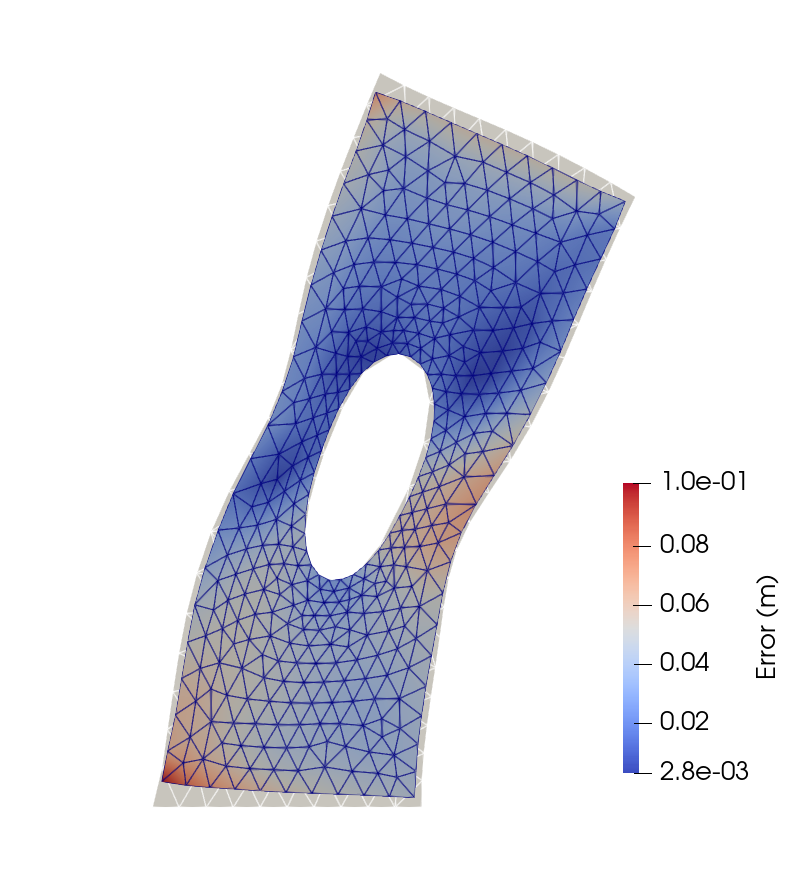}
  \caption{VBI-FEM result}
\end{subfigure}
\caption{The inverse recovery results of the deformation mapping ${ \psi} ({\bf u} ({\bf X}))$
for a plate with a hole in case II.}
\label{fig:Example 1 disp recovery process case 2}
\end{figure}

Then, the same recovery processes are performed by using
two other non-rigid point registration methods, i.e.
the coherent point drift (CPD) \cite{myronenko2010point}
and the preserving global and local structures (PR-GLS) method \cite{ma2015non}.
The hyperparameters are tuned to achieve the optimal results.
The comparisons of the displacement recovery accuracy among two cases are shown
in Table \ref{table: example 1 comparison of accuracy between three cases}.

As shown in Table \ref{table: example 1 comparison of accuracy between three cases},
all three methods have good accuracies for the case I,
the recovery result accuracy of VBI-FEM is between that of CPD and PR-GLS.

Similarly, the convergence and accuracy of the displacement recovered for case II
are also performed.
In this case, the proposed VBI-FEM has the best accuracy.
As one can see that the VBI-FEM method has high accuracies in displacement recovery in both cases,
even though different meshes are used to represent the deformed configurations
and different material models are chosen in the recovery process.

\begin{table}[ht]
\centering
\caption{The comparison of displacement recovery error between different methods.}
\label{table: example 1 comparison of accuracy between three cases}
\begin{tabular}{cccc}
\hline \bf Recovery error [$\%$] & \bf CPD & \bf PR-GLS & \bf VBI-FEM \\
\hline
\bf Case I & $7.4$  & $4.1$ & $5.3$ \\
\bf Case II & $10.2$  & $8.8$ & $7.6$ \\
\hline
\end{tabular}
\end{table}

\begin{figure}[ht]
\centering
\begin{minipage}{0.49\linewidth}
\begin{center}
  \includegraphics[width=2.0in]{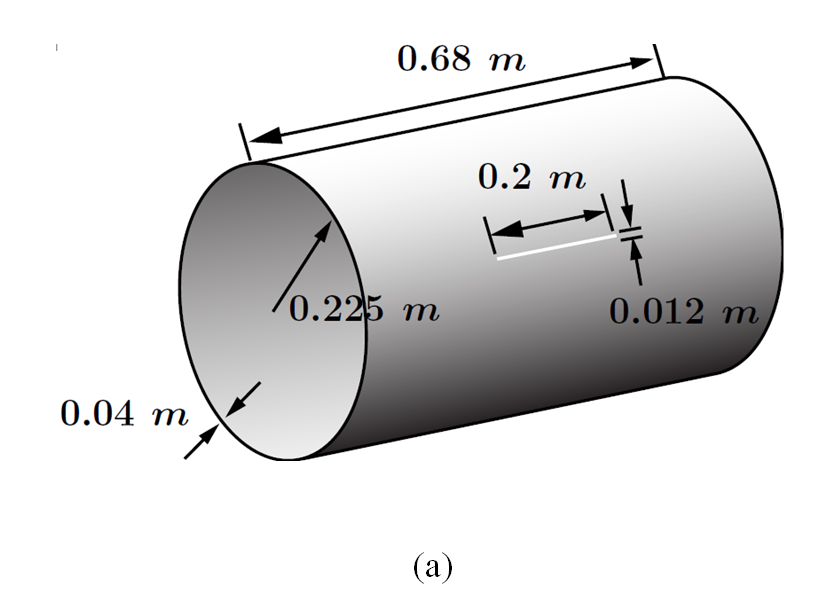}
\end{center}
\end{minipage}
\centering
\begin{minipage}{0.70\linewidth}
\begin{center}
  \includegraphics[width=3.0in]{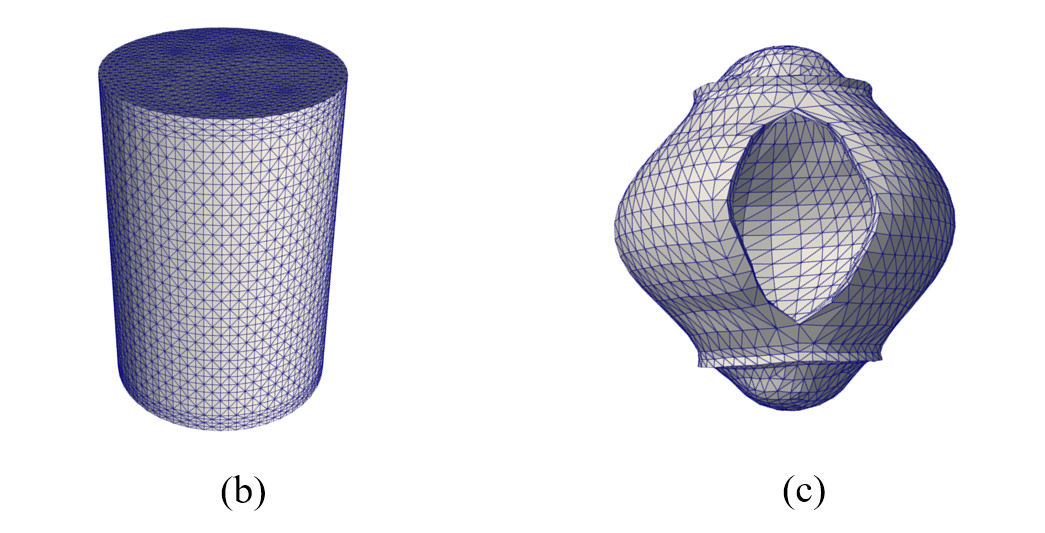}
\end{center}
\end{minipage}
\caption{ The finite element meshes for the thin cylindrical shell:
(a) The setup of the thin shell cylinder;
(b) A finer mesh in the initial configuration $\Omega_X$;
(c) Mesh in the final configuration $\Omega_x$.}
\label{fig:cylinder meshes}
\end{figure}

\subsection{Solid cylinder subjected to internal pressure}
\label{section: cyliner example}
In this subsection, we apply the VBI-FEM method to recover the displacement
field of a solid thin shell cylinder subjected to internal pressure.
The dimension of the cylinder is shown in Fig. \ref{fig:cylinder meshes} (a).
The cylinder is modeled by the Gurson-Tvergaard-Needleman(GTN)
material constitutive model \cite{Springmann2008}.
The thin shell cylinder is subjected to an internal pressure
of $10~ MPa$. A detailed numerical simulation of the cylinder
was reported in \cite{qian2008meshfree}.
In this numerical example, we use VBI-FEM method to recover
the deformation of the solid thin cylinder based on the initial and deformed
configurations reported in \cite{qian2008meshfree}.

The FE model for this solid thin cylinder has 13,848 three-dimensional tetrahedron elements
in deformed configuration, which is shown in Fig.  \ref{fig:cylinder meshes}(c).
A background mesh is generated on the initial configuration with 110,784 tetrahedron elements,
which is shown in Fig. \ref{fig:cylinder meshes}(b).
Under the internal pressure, a crack will grow along the initial notch in the solid cylinder,
the simulated crack growth is shown in Fig. \ref{fig:cylinder meshes}(c).
For this example, the mesh
in the current configuration is set as the GMM centroids,
and the dense mesh in the initial configuration is chosen as the data points.
Therefore, during inverse solution recovery process,
the deformed mesh shown moves back to the initial configuration
in order to align with the undeformed dense mesh.
Furthermore, in contrast to the original forward numerical simulation
in which the Gurson-Tvergaard-Needleman (GTN)
model is used as the material constitutive model \cite{GTN84},
a linear elastic constitutive model is chosen as the prior elastic potential in the VBI-FEM solution.
The values of the elastic constants and other inverse solution
model parameters are given as follows:
Lam\'e's constants $\lambda = 0.00 1 MP_a$ and $\mu = 0.001 MP_a$; the variational Bayesian learning
hyper-parameters are: $\beta = \num{4 e-4}$ and $\gamma =\num{1e-5}$, and maximum iteration number $n= 50$.

\begin{table}[ht]
\centering
\caption{The comparison of displacement recovery error between different methods.}
\label{table: example 2 comparison of accuracy between three cases}
\begin{tabular}{cccc}
\hline \bf Recovery error [$\%$] & \bf CPD & \bf PR-GLS & \bf VBI-FEM \\
\hline
\bf Thin shell cylinder & $22.7$  & $134.5$ & $10.9$ \\
\hline
\end{tabular}
\end{table}

\begin{figure}[ht]
\centering
  \includegraphics[width=\columnwidth]{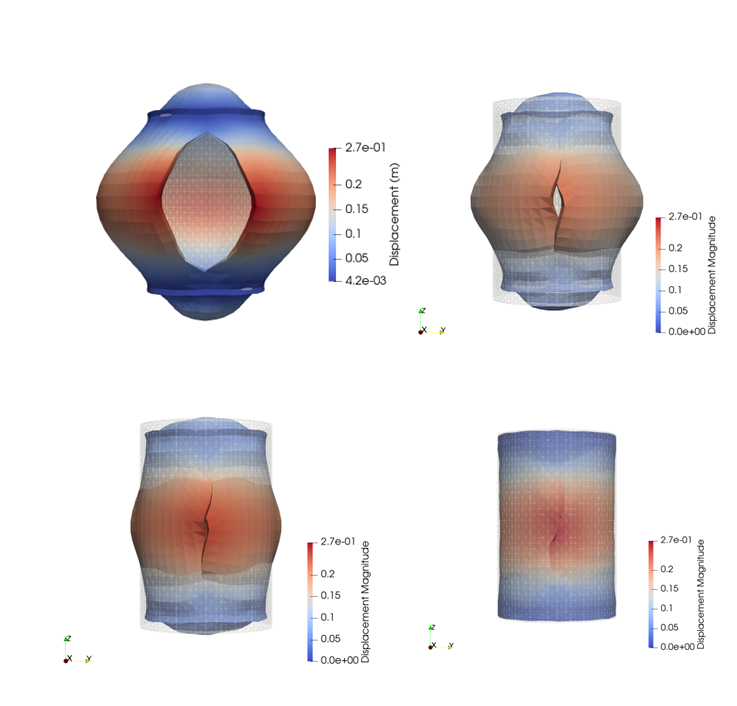}
\caption{The inverse recovery results for the deformation mapping of a three-dimensional
fractured cylinder: ${\bfg \psi}^{-1} ({\bf u} ({\bf X}))$.}
\label{fig:Fig8}
\end{figure}

Figure \ref{fig:Fig8} displays the displacement recovery process for the cracked
solid thin cylinder.
Similarly, the background mesh in the reference configuration is shown
in gray as the true inverse deformation mapping range.
The current configuration mesh is color-coded by the magnitude of error
reference to the initial configuration through the inverse solution process.
After 30 steps of iterations in the inverse solution,
the average nodal inverse displacement error, which is calculated by Eq. (\ref{eq:error-calculation}),
is decreased from $0.11 ~m$ to $0.012~ m$.
That indicates that $89.1\%$ of the total inverse displacement is recovered by the VBI-FEM method.
This example clearly demonstrates that the proposed VBI-FEM
is capable of the recovery of deformation mapping
of the three-dimensional structures with defects.

\begin{figure}[ht]
\centering
\begin{subfigure}{.32\columnwidth}
  \centering
  \includegraphics[width=\columnwidth]{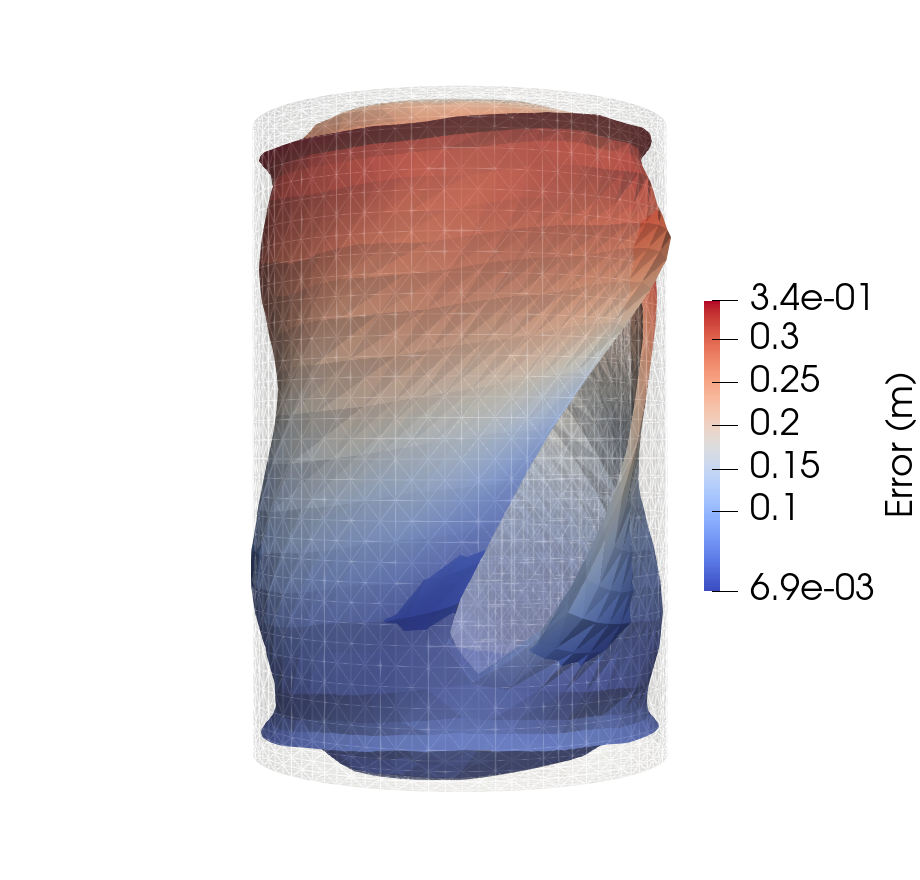}
  \caption{PR-GLS result}
\end{subfigure}
\begin{subfigure}{.32\columnwidth}
  \centering
  \includegraphics[width=\columnwidth]{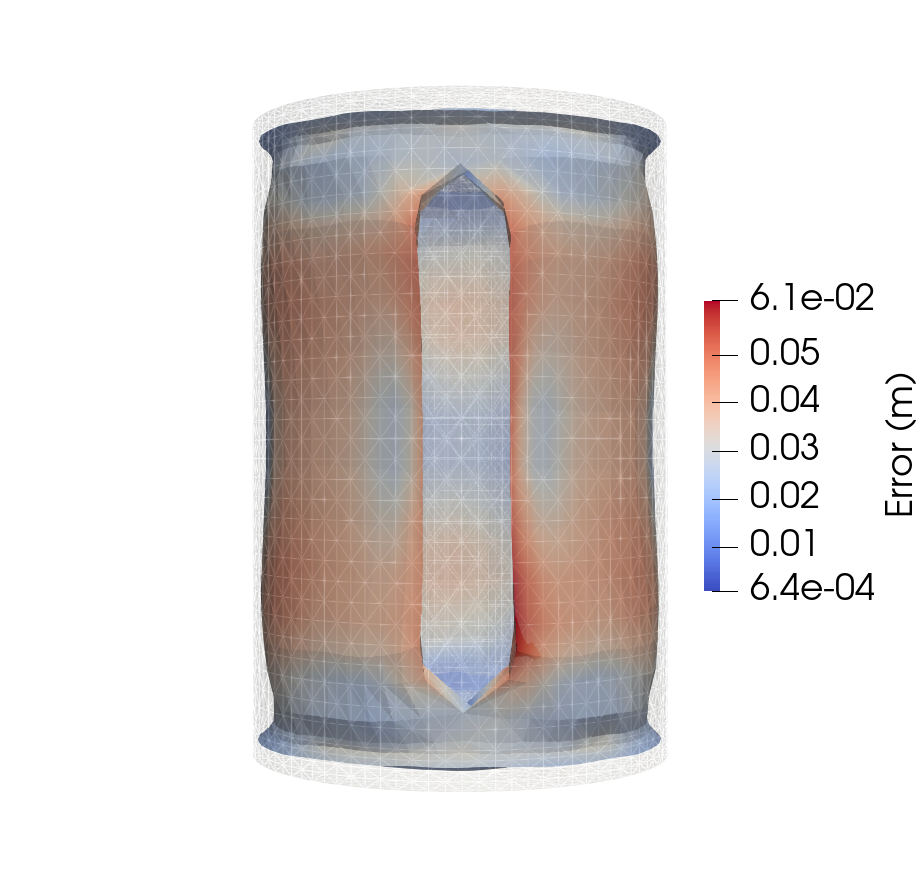}
  \caption{CPD result}
\end{subfigure}
\begin{subfigure}{.32\columnwidth}
  \centering
  \includegraphics[width=\columnwidth]{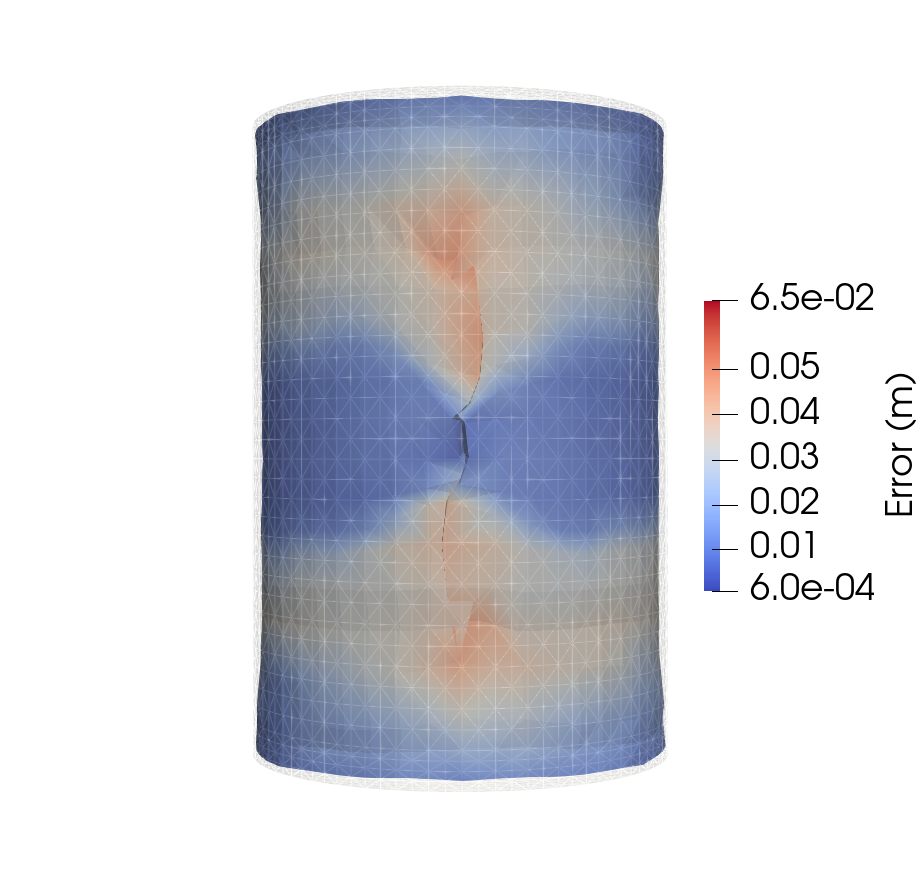}
  \caption{VBL-FEM result}
\end{subfigure}
\caption{Comparison of the inverse recovery results for a 3D
fractured cylinder for three different methods.}
\label{fig:Fig9}
\end{figure}

In Fig. \ref{fig:Fig9}, we compare the final results or the final configurations of the
inverse recovery solutions of three different methods:
PR-GLS, CPD, and VBI-FEM.
As shown in Fig. \ref{fig:Fig9}, only the VBI-FEM solution
can successfully recover three-dimensional deformation pattern with a defect,
while both PR-GLS and CPD method only recover the initial configuration
layout but not point-wise deformation field. Therefore, both methods
cannot close the crack, which clearly demonstrates
the VBL-FEM method's potential to become a robust and efficient inverse
solution to recovery complex three-dimensional deformation mappings.

\section{Summary}
\label{sec: Conclusion}
In this work,
a mixed variational Bayesian inference finite element method has been
developed, which
provides a machine intelligent numerical solution for
solving the inverse problem of
recovering continuum deformation mappings.
In this inverse solution, we only know the layout of the initial undeformed
and final deformed boundary configurations of continuum solids,
and the deformed structures may contain defects and strong discontinuities.
In parallel with the well-known E-M algorithm in statistics, we developed
a staggered computational algorithm for the proposed VBI-FEM formulation,
which is termed as the FE-BL (Finite Element/Bayesian Learning) operator splitting
algorithm.

Compared with other non-rigid registration algorithms in computer vision and image diagnosis,
the proposed VBI-FEM approach is an inverse solution for recovering deformation mapping in any dimension ($n \ge 1$)
including but not limited to 3D bulk continuum, whereas the state-of-the-art GMM-based point
registration algorithms are only valid for matching the surface or domain boundary points in principle.
Moreover,  the VBI-FEM inverse solution does not require the data point sets
in the initial and final configurations having a one-to-one correspondence,
in other words, VBI-FEM allows the two point sets to have different
number of points that have no correlation.
Furthermore, the point registration performance of the VBI-FEM
is insensitive to the values of hyperparameters used in VBI-FEM, i.e., $\beta$ and $\gamma$,
which may vary in a range of several orders of magnitudes without affecting inverse  solution
results.

It is demonstrated in this paper that the VBI-FEM can provide
practically useful inverse solutions with good accuracy,
and this inverse finite element solution procedure can be applied to a host
of engineering problems, such as structural forensic analysis,
digital image correlation, predictive modeling of geometric deviations
in 3D printing, AI-aided diagnosis of medical images, and inverse solutions
for the partial differential equations in general.

We would like to emphasize that
in the VBI-FEM inverse solution we assume that boundary of the elastic solids
can only have constant or uniform strain fields.
For a given boundary contour shape, it may correspond to both
uniform boundary strain field and the non-uniform strain fields.
In these cases, the VBI-FEM inverse solution for the deformation maps
cannot be unique, or the proposed method cannot distinguish their differences.
Nevertheless, the proposed method may still be applied
to solve the cases where the inverse solution is not unique.
In those cases, we believed that
the numerical solution of the VBI-FEM may be the one that
has the largest probability to be captured.
\color{black}

\medskip
\section*{Acknowledgements}
The authors would like to thank the anonymous reviewers for their
comments and suggestions that improve the quality of the paper.

\medskip
{\appendix[Proof of Theorem \ref{theorem: unique solution}]
\begin{IEEEproof}
The stationary condition for attaining the extreme of
$Q ({\tilde{\bf u}}, \Sigma; {\tilde{\bf u}}_n, \Sigma_n)$ is as follows,
\begin{eqnarray}
&&\delta Q ({\tilde{\bf u}},\Sigma; {\tilde{\bf u}}_n, \Sigma_n)=
\delta_{\Sigma} Q({\tilde{\bf u}},\Sigma
; {\tilde{\bf u}}_n, \Sigma_n) +
\delta_{{\bf u}}
Q({\tilde{\bf u}},\Sigma ; {\tilde{\bf u}}_n, \Sigma_n) = 0 ,
\nonumber
\\
&&~~~~~~~~~~~~\forall (\delta {\tilde{\bf u}}, \delta \Sigma) \in
\mathcal{V}_0 \oplus \mathcal{R}^+
\label{eq:weak-form}
\end{eqnarray}
where
\begin{eqnarray}
    &&\delta_{\Sigma} Q ({\tilde{\bf u}},\Sigma; {\tilde{\bf u}}_n, \Sigma_n) =
    \int_{\Omega_x}\int_{\Omega_X} p\left(\mathbf{X+{\tilde{\bf u}}_{n}} | \mathbf{x}\right) \cdot
    \nonumber
    \\
 &&   \left(
 -{\frac{\left\| {\bf X}+ {\tilde{\bf u}} - {\bf x} \right\| ^2}{\Sigma^{3}}}+
    \frac{D}{\Sigma}\right)
    \cdot \delta \Sigma
    d\Omega_X d\Omega_x
    \label{eq:weak-form1}
\end{eqnarray}
and
\begin{eqnarray}
&&
\delta_{{\tilde{\bf u}}} Q ({\tilde{\bf u}},\Sigma; {\tilde{\bf u}}_n, \Sigma_n)
\nonumber
\\
&=& \beta \int_{\Omega_x}\int_{\Omega_X}
p\left(\mathbf{X+{\tilde{\bf u}}_n} | \mathbf{x}\right) \cdot
{\Sigma}({\bfg \varepsilon}({\tilde{\bf u}})):\delta { \varepsilon}
d\Omega_X d\Omega_x
\nonumber
\\
&+&
\frac{1}{\Sigma^2}\int_{\Omega_x}\int_{\Omega_X}
p\left(\mathbf{X+{\tilde{\bf u}}_{n}} | \mathbf{x}\right)\cdot
(\mathbf{X}+ {\tilde{\bf u}}-\mathbf{x})\delta
{\tilde{\bf u}}
d\Omega_X d\Omega_x
\nonumber
\\
 &+& \gamma \int_{\Omega_{X}} \delta \phi({\tilde{\bf u}}) d \Omega_{X}
\label{eq:weak form2}
\end{eqnarray}
Since $\delta \Sigma$ and $\delta {\tilde{\bf u}}$ are independent, we have
\begin{equation}
\delta Q ({\tilde{\bf u}},\Sigma; {\tilde{\bf u}}_n, \Sigma_n) =0 ~\to
\left\{
\begin{array}{rcl}
\delta_{{\tilde{\bf u}}}Q ({\tilde{\bf u}},\Sigma; {\tilde{\bf u}}_n, \Sigma_n) =0
\\
\\
\delta_{\Sigma} Q ({\tilde{\bf u}},\Sigma; {\tilde{\bf u}}_n, \Sigma_n) =0
\end{array}
\right.
\end{equation}
From $\delta_{\Sigma} Q ({\tilde{\bf u}},\Sigma; {\tilde{\bf u}}_n, \Sigma_n)  =0$
and using Eq. (\ref{eq:weak-form1}), we can obtain
\begin{equation}
\Sigma^2 = {\frac{1}{D}}
   \int_{\Omega_x}\int_{\Omega_X}
     p\left(\mathbf{X}+{\tilde{\bf u}}_{n} | \mathbf{x}\right)
 \left\| {\bf X}+ {\tilde{\bf u}} - {\bf x} \right\|^2
    d\Omega_X d\Omega_x
    \label{eq:expectation1}
\end{equation}

The sufficient condition for $Q({\tilde{\bf u}},\Sigma^2; {\tilde{\bf u}}_n, \Sigma_n )$
having an infimum is:
\begin{eqnarray}
\delta^2 Q({\tilde{\bf u}},\Sigma^2; {\tilde{\bf u}}_n, \Sigma_n) &=&
\delta^2_{{\tilde{\bf u}} {\tilde{\bf u}}} Q({\tilde{\bf u}},\Sigma; {\tilde{\bf u}}_n, \Sigma_n)
\nonumber
\\
&+& 2\delta^2_{{\tilde{\bf u}} \Sigma}  Q({\tilde{\bf u}},\Sigma; {\tilde{\bf u}}_n, \Sigma_n)
\nonumber
\\
&+& \delta^2_{\Sigma \Sigma} Q({\tilde{\bf u}},\Sigma; {\tilde{\bf u}}_n, \Sigma_n)
\nonumber
 \\
&>& 0~,
\label{eq:SupremeC}
\end{eqnarray}
where
\begin{eqnarray}
\delta^2_{{\tilde{\bf u}} {\tilde{\bf u}}} Q
&=& \int_{\Omega_x}\int_{\Omega_X}
p\left(\mathbf{X}+ {\tilde{\bf u}}_{n} | \mathbf{x}\right) \cdot
\Bigl(
\delta { \varepsilon}:
\gamma {\frac{\partial^2 W}{\partial {\varepsilon} \partial { \varepsilon}}}
:\delta { \varepsilon}
+{\frac{1}{\Sigma^{2}}} \delta {\tilde{\bf u}} \cdot \delta {\tilde{\bf u}}
\Bigr)
\nonumber
 \\
&&
d\Omega_X d\Omega_x
+ \gamma \int_{\Omega_{X}} \delta {\tilde{\bf u}} \cdot {\partial^2 \phi \over \partial {\tilde{\bf u}}
\partial {\tilde{\bf u}}} \cdot \delta {\tilde{\bf u}} d \Omega_X
\end{eqnarray}
\begin{eqnarray}
\delta^2_{\Sigma \Sigma} Q &=&
\int_{\Omega_x}\int_{\Omega_X} p\left(\mathbf{X}+ {\tilde{\bf u}}_n | \mathbf{x}\right) \cdot
\left(
\frac{3 \left\| {\bf X}+ {\tilde{\bf u}} - {\bf x} \right\|^2}{\Sigma^4}
-  \frac{D}{\Sigma^2}
\right)
\nonumber
\\
&& \cdot (\delta \Sigma)^2
d\Omega_X d\Omega_x
\end{eqnarray}
\begin{eqnarray}
\delta^2_{\Sigma u} Q =
-\int_{\Omega_x}\int_{\Omega_X} p\left(\mathbf{X}+ {\tilde{\bf u}}_{n} | \mathbf{x}\right) \cdot \left(
{\frac{ ({\bf X}+ {\tilde{\bf u}} - {\bf x} )}{ \Sigma^{3}}}\right)\cdot
\delta {\tilde{\bf u}} \delta \Sigma
d\Omega_X d\Omega_x
\end{eqnarray}

We can rewrite $\delta^2 Q ({\tilde{\bf u}}, \Sigma; {\tilde{\bf u}}_n, \Sigma_n)$ as
\begin{eqnarray}
&&\delta^2  Q ({\tilde{\bf u}}, \Sigma; {\tilde{\bf u}}_n, \Sigma_n)
=
\int_{\Omega_x}\int_{\Omega_X}
p\left(\mathbf{X}+{\tilde{\bf u}}_{n} | \mathbf{x}\right) \cdot
\nonumber
\\
&&\Bigl(
\delta { \varepsilon}:
\gamma {\partial^2 W \over \partial { \varepsilon} \partial { \varepsilon}}
:\delta { \varepsilon}  \Bigr)
d\Omega_X d\Omega_x
 + \gamma \int_{\Omega_{X}}
\delta {\tilde{\bf u}} \cdot {\partial^2 \phi \over \partial {\tilde{\bf u}}
\partial {\tilde{\bf u}}} \cdot \delta {\tilde{\bf u}} d \Omega_{X}
\nonumber
\\
&&+{1 \over \Sigma^{2}}
\int_{\Omega_x}\int_{\Omega_X}
p\left(\mathbf{X}+ {\tilde{\bf u}}_{n} | \mathbf{x}\right)
\Bigl(
\delta {\tilde{\bf u}} -
\frac{{\bf X}+ {\tilde{\bf u}} - {\bf x}}{\Sigma^{2}} \delta \Sigma
\Bigr)^2
d\Omega_X d\Omega_x
\nonumber
\\
&&+
\int_{\Omega_x}\int_{\Omega_X}
p\left(\mathbf{X}+ {\tilde{\bf u}}_{n} | \mathbf{x}\right)
\left(
\frac{2 \left\| {\bf X}+ {\tilde{\bf u}} - {\bf x} \right\|^2}{\Sigma^4}
-  \frac{D}{\Sigma^2}
\right)
\nonumber
\\
&&
\cdot
(\delta \Sigma)^2
  d\Omega_X d\Omega_x~.
 \label{eq:second-order-term}
\end{eqnarray}

Because the regularization term is positive definite,
the first term in Eq. (\ref{eq:second-order-term})
is positive (definite), if the strain energy is convex.
The second term of Eq. (\ref{eq:second-order-term})
is always positive.
Considering Eq. (\ref{eq:expectation1}), we can rewrite
the last term of Eq. (\ref{eq:second-order-term}) and
show that
\begin{eqnarray}
   && \int_{\Omega_x}\int_{\Omega_X}
    p\left(\mathbf{X}+ {\tilde{\bf u}}_n | \mathbf{x}\right)
    \left(
  \frac{2 \left\| {\bf X}+ {\tilde{\bf u}} - {\bf x} \right\|^2}{\Sigma^4}
  -  \frac{D}{\Sigma^2}
    \right)
    \nonumber
    \\
    &&\cdot
    (\delta \Sigma)^2
      d\Omega_X d\Omega_x~
      =\Bigl( {2 \Sigma^2 D \over \Sigma^4} - {D \over \Sigma^2} \Bigr)(\delta \Sigma)^2
      = D{ (\delta \Sigma)^2 \over \Sigma^2} > 0~.
      \label{eq:second-order-term2}
\end{eqnarray}

Thus, we can conclude that
\[
\delta^2 Q ({\tilde{\bf u}}, \Sigma; {\tilde{\bf u}}_n, \Sigma_n) >0~.
\]
\medskip

This confirms that the mixed probabilistic variational principle
(Eq.(\ref{eq:weak-form})) has a global minimizer, i.e.
there exits a solution
$({\bf u}^{\star}, \Sigma^{\star}) \in \mathcal{V} \oplus \mathcal{R}^+$
such that it minimizes $Q ({\tilde{\bf u}}, \Sigma; {\tilde{\bf u}}_n, \Sigma_n)$,
\[
Q({\bf u}^{\star}, \Sigma^{\star}; {\tilde{\bf u}}_n, \Sigma_n) =
\inf\limits_{({\tilde{\bf u}}, \Sigma) \in \mathcal{V} \oplus \mathbb{R}^+}
Q({\tilde{\bf u}}, \Sigma; {\tilde{\bf u}}_n, \Sigma_n)~.
\]
\end{IEEEproof}



\ifCLASSOPTIONcaptionsoff
  \newpage
\fi


\bibliography{main}
\bibliographystyle{plain}{}

\begin{IEEEbiography}[{\includegraphics[width=0.7in,height=1.0in,clip,keepaspectratio]{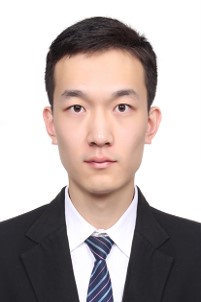}}]{Chao Wang}
received the B.S. degree from Harbin Engineering University in 2017 with a major in Ocean Engineering 
and the Ph.D. degree in Civil Engineering from the University of California
at Berkeley in 2022.
He is currently working as an R\&D Engineer at ANSYS.
His research area is developing artificial intelligence and machine
learning-based methods to solve engineering problems, including computer vision
and Bayesian neural networks.
\end{IEEEbiography}

\vspace{11pt}

\begin{IEEEbiography}[{\includegraphics[width=0.7in,height=1.0in,clip,keepaspectratio]{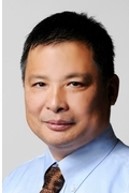}}]{Shaofan Li}
received the B.S. degree in Mechanical Engineering from the East China University of Science and Technology in 1982
and M.S. degrees from the Huazhong University of Science in 1989 and Technology and the University of Florida in 1993 with majors in Computational Mechanics and Aerospace Engineering, respectively.
Then he received the Ph.D. degree in Mechanical Engineering from Northwestern University in 1997.
Dr. Shaofan Li is currently a full professor at the University of California at
Berkeley. He has been working on engineering applications of artificial intelligence and machine learning neural 
networks.
\end{IEEEbiography}

\vfill

\end{document}